\documentclass[10pt,twocolumn,letterpaper]{article}

\usepackage[pagenumbers]{cvpr}
\usepackage{booktabs}
\usepackage{xcolor}
\usepackage{multirow}
\usepackage{makecell}
\usepackage{colortbl}
\usepackage{adjustbox}
\usepackage{xcolor}
\usepackage[normalem]{ulem}

\usepackage{algorithm}
\usepackage{algpseudocode}
\usepackage{graphicx}
\usepackage{xcolor}
\usepackage{amssymb}
\newcommand{\stopgrad}{\operatorname{sg}}
\usepackage{tabularx}
\usepackage{dblfloatfix}

\definecolor{oursbg}{gray}{0.92}
\definecolor{gaingreen}{RGB}{0,128,64}
\definecolor{oraclegray}{gray}{0.5}
\newcommand{\gain}[1]{\textsubscript{\,\textcolor{gaingreen}{#1}}}

\definecolor{cvprblue}{rgb}{0.21,0.49,0.74}
\usepackage[pagebackref,breaklinks,colorlinks,allcolors=cvprblue]{hyperref}

\def\confName{CVPR}
\def\confYear{2027}

\title{InCue: Making Better Use of What Multimodal LLMs Already See at Test Time}

\author{Shuwei Wu$^{1}$\qquad Siyue Yu$^{2}$\qquad Longin Jan Latecki$^{1}$\thanks{Corresponding author.}\\
$^{1}$Temple University\qquad $^{2}$Xi'an Jiaotong-Liverpool University\\
{\tt\small \{shuwei.wu, latecki\}@temple.edu, siyue.yu02@xjtlu.edu.cn}
}

\begin{document}
\maketitle
\begin{abstract}
Even when multimodal large language models (MLLMs) answer incorrectly, their internal
cues often point to the relevant region. A widely used remedy re-presents the
cue-selected region as an enlarged crop to supply finer detail. However, we find that much of
its gain requires no new detail: it largely survives when the crop comes from the
already-seen low-resolution input, and partly even when the region's cached tokens are
merely copied. Part of re-presentation's gain thus comes from drawing attention to evidence the model already holds. We propose \textbf{InCue}, a test-time framework that makes better use of this evidence without additional visual input. InCue treats visual
attention and predictive-entropy gradients as \textbf{In}ternal \textbf{Cue}s, couples
them into a spatial target, and optimizes only an offset on the existing visual tokens to
steer attention toward it. Under matched visual-token budgets, InCue achieves state-of-the-art performance, consistently outperforming recent test-time methods and improving four MLLMs by up to
8.9\% on V$^*$ and 23.4\% on its hard samples. Its gains carry over to eight
general benchmarks. Code will be available at \url{https://github.com/paul-swu/InCue}.
\end{abstract}

\section{Introduction}
\label{sec:intro}
A growing body of work observes that even when multimodal large language models (MLLMs) answer incorrectly, their internal cues, such as attention and gradient maps, may still point to the relevant region~\cite{zhang2025vicrop,liu2026aif}. However, pointing to the region only reveals where it is, whereas answering requires recognizing what it contains. Since the evidence is often small, its features at the limited input resolution of the vision encoder lack the fine-grained details needed for such recognition~\cite{zhang2025vicrop}. Accordingly, a widely used family of inference-time methods~\cite{gropl2026egg,zhong2025focus,xie2026havc,yi2026hiperson,zhang2025vicrop,ma2026slofo} adopts these cues to select the indicated region, crops and enlarges it from the original image, and re-presents it as an additional input to supply finer details.
\begin{figure}[t]
  \centering
  \includegraphics[width=\linewidth]{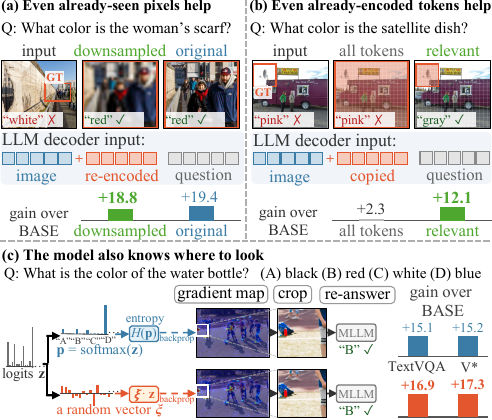}
\caption{\textbf{Evidence is already in the model.}
    (a):~Re-encoding the downsampled GT region helps; the original adds only 0.5\% (V$^*$, Qwen2.5-VL-7B).
    (b):~Copying the region's cached tokens also helps (Visual Genome~\cite{krishna2017visual}, LLaVA-1.5-7B~\cite{liu2024llava15}). Orange in (a, b): re-fed region and tokens.
    (c):~Replacing EGG's entropy (top) with a random projection (bottom) selects crops as well (TextVQA~\cite{singh2019textvqa}, LLaVA-1.5-7B; V$^*$, full EGG, LLaVA-1.6-Mistral-7B~\cite{liu2024llava16}).}
  \label{fig:diag}
\end{figure}

However, re-presentation adds visual tokens to the input, and the crop isolates the region from the context some questions depend on. To investigate whether it is necessary, we first examine the source of its gain. If it came from the finer details, enlarging the same region from the downsampled image the model actually receives should help little, as upsampling only interpolates existing pixels. We therefore crop the ground-truth region from both the high-resolution original and the downsampled input, re-encode the two crops at the same resolution, and append their tokens to the image tokens. Both crops bring large gains on V$^*$~\cite{wu2024vstar} (Fig.~\ref{fig:diag}a), with the original crop only 0.5\% above the downsampled one. Much of the gain of re-presentation thus requires no new detail.

Yet enlarging and re-encoding still change the region's representation. We therefore examine whether re-encoding is necessary by appending a copy of its cached tokens to the input, adding no new details or features. This alone improves accuracy by 12.1\% (Fig.~\ref{fig:diag}b), whereas copying as many tokens from the whole image gives only 2.3\%. The evidence is thus already encoded in the model's representations, and re-presentation also helps the model use it, by giving the relevant tokens a larger share of the LLM's attention; raising their attention logits instead gives a similar gain (Supp.). Part of re-presentation's gain thus needs no additional visual input, which motivates steering the attention directly toward the relevant tokens.

\begin{figure}[t]
  \centering
  \includegraphics[width=\columnwidth]{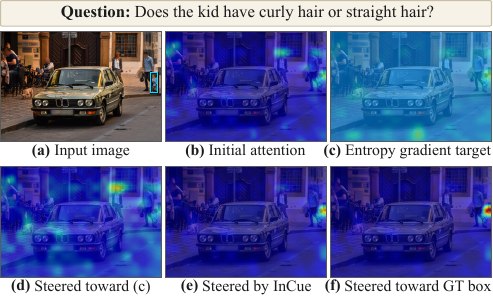}
\caption{\textbf{Steering attention toward different targets.}
    Cyan box: queried kid (reference only).
    (c)~is near-uniform; (d):~~fitted to (c) with a KL loss, the attention flattens and misses the kid. (e):~InCue couples (b) and (c) to reinforce where both are strong, focusing on the kid, close to the GT-box oracle (f). Using Qwen2.5-VL-7B~\cite{bai2025qwen25vl}.
    }
  \label{fig:steer}
\end{figure}

Steering the attention requires a spatial target; a natural source is the internal cues existing methods use to select regions. Among them, Entropy-Gradient Grounding (EGG)~\cite{gropl2026egg} backpropagates the predictive entropy, a scalar measure of uncertainty, to reveal its sensitivity to each position. To test whether selection needs this answer-specific cue, we keep EGG's pipeline and replace only the entropy with a projection of the output logits onto a fixed random vector free of answer semantics. Surprisingly, this random readout selects regions as well as the entropy does (Fig.~\ref{fig:diag}c). Useful localization is thus not unique to the entropy, and region selection needs only a coarse location: finding the region is not the hard part. Yet the methods above use these cues only to pick a region to re-present, leaving the answer-specific information in the cue underutilized. We therefore ask: \textit{Can an MLLM's own internal cues guide it to make better use of its already-encoded visual representations, without additional visual inputs or external prompts?}

To explore this possibility, a natural first attempt is to steer the attention directly toward the entropy gradient~\cite{simonyan2014deep,selvaraju2017gradcam} by optimizing the existing visual representations~\cite{yang2025lra,xing2026activationreplay,wu2024controlmllm,wu2026controlmllmpp}. However, a cue that suffices for region selection is not necessarily a good target for steering the attention: selection only needs it to peak at the right place, whereas fitting imposes its entire shape on the attention. As this map is almost uniform (Fig.~\ref{fig:steer}c), the steered attention flattens and disperses (Fig.~\ref{fig:steer}d), and the answers barely improve. Locating the region is easy, but turning the cues into a spatial target that the attention can follow is not.

\begin{figure*}[t]
    \centering
    \includegraphics[width=\textwidth]{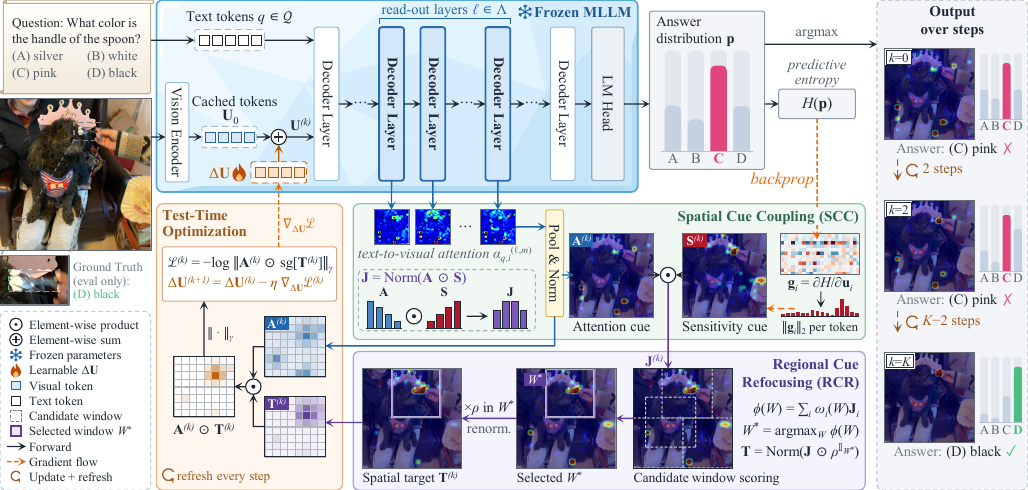}
    \caption{\textbf{Overview of InCue.} The MLLM stays frozen, and the only free variable is an
additive offset $\Delta U$ on the cached visual tokens $U_0$. SCC couples the read-out
attention cue $A$ with the sensitivity cue $S$, obtained by backpropagating the predictive
entropy $H(p)$, into the joint response $J$. RCR amplifies $J$ inside the best-scoring window
$W^\star$ to form the spatial target $T$. Test-time optimization then updates $\Delta U$ to
increase the overlap between the current attention $A^{(k)}$ and $T^{(k)}$. The cues and the
target are refreshed at every step $k$, and the answer is read out after $K$ steps (Output):
here it changes from (C) pink to the correct (D) black.}
    \label{fig:main_incue}
\end{figure*}

To obtain a target that the attention can follow, we pair the entropy gradient with the model's attention: the former reflects where the prediction is sensitive, the latter where the model currently looks. Both respond to the relevant region, yet contain noise from different sources (Fig.~\ref{fig:steer}b--c). Starting from where the model already looks keeps the target within reach, while the sensitivity reinforces the positions where both cues respond and holds in check those that stand out in only one. A target built on the cues' joint response concentrates the steered attention on the relevant region and suppresses off-target responses from either source (Fig.~\ref{fig:steer}e). Building on these insights, we propose \textbf{InCue}, a test-time framework that steers an MLLM's attention over its existing visual tokens toward a spatial target built from its own internal cues. Its Spatial Cue Coupling (SCC) module computes this joint response, turning the sensitivity from a cue for region selection into a target for steering the attention.

However, the joint response is defined over individual visual positions and may still be scattered, whereas the evidence usually occupies a contiguous region. We therefore design a Regional Cue Refocusing (RCR) module, following prior window-based selection~\cite{zhang2025vicrop,gropl2026egg}: it scans candidate windows over the joint response, selects the most responsive one, and amplifies the positions inside it, turning position-level response into region-level guidance while keeping the differences within the window.

With the spatial target in place, InCue steers the attention toward it by optimizing a learnable offset on the visual tokens, with the model frozen. Since each update changes the attention, prediction, and sensitivity, both cues and the target are refreshed at every step. Finally, the answer is generated from the adjusted tokens, so InCue makes better use of the visual tokens the model already has without any additional visual input. Our contributions are as follows:

\textbullet \ We show that much of re-presentation's gain needs no new visual detail, and propose InCue, which steers the attention toward a target built from the model's internal cues.

\textbullet \ We propose SCC to couple the model's attention with the spatial sensitivity of its predictive entropy. Although the two cues differ, tokens scoring high under both tend to carry the key visual evidence for answering, which motivates steering the attention toward their overlap.

\textbullet \ We design RCR to refocus the target on the high-response region, providing regional guidance.

\textbullet \ Under matched visual-token budgets, InCue achieves state-of-the-art (SOTA) performance, improving four MLLMs by up to 8.9\% on V$^*$ and 23.4\% on its hard samples; its gains carry over to eight general benchmarks.

\section{Related Work}
\label{sec:related}

\paragraph{Fine-grained perception.}
Small objects lose detail at a limited input resolution.
One line of work therefore gives the model more visual input, through higher-resolution or tiled encoding~\cite{liu2024llava16,chen2024internvl15,wang2024qwen2vl} or by training the model to search for or zoom into key regions~\cite{wu2024vstar,shao2024visualcot,man2025argus,zheng2026deepeyes,lai2026minio3,feng2026blink}.
Training-free methods instead re-present the question-relevant region at inference, as an enlarged crop appended to the input or as a re-rendered image that highlights it~\cite{dalal2026attwarp,yu2024api,liu2026sbnb}.
To find the relevant content, some search, retrieve, or analyze sub-images~\cite{wang2024hrbench,shen2025zoomeye,li2025dyfo,wang2025rap,kim2026ug}; others select the region with the model's internal cues, such as attention~\cite{zhang2025vicrop,zhong2025focus,xie2026havc,liu2026hide}, the entropy gradient~\cite{gropl2026egg}, or attention combined with gradient-based relevance maps~\cite{yi2026hiperson,ma2026slofo}.
Several studies also analyze what limits the fine-grained perception of MLLMs~\cite{zhang2026limits,liu2026hide,wei2026zwz}.
Our diagnostics (Fig.~\ref{fig:diag}) show that much of re-presentation's gain requires no new detail, and part of it comes from drawing attention to evidence the model already holds. InCue uses the internal cues to build a spatial target for attention rather than to select a region to re-present: it steers the attention over the existing visual tokens without additional visual input, and it complements re-presentation (Tab.~\ref{tab:crop}).

\paragraph{Test-time methods.}
Instead of changing the input, another line of work changes how the model uses it at test time.
Since image tokens receive little attention in the LLM~\cite{chen2024fastv}, the most direct approach is to adjust attention, hidden states, or decoding with hand-designed rules~\cite{liu2024pai,kang2025var,chen2025adaptvis,liu2026aif,peng2026vrga,zhao2026vga,zou2025memvr,liu2025vti,leng2024vcd,huang2024opera}.
Other methods update parameters or prompts, by minimizing the predictive entropy~\cite{wang2021tent,shu2022tpt} or by reinforcement learning that rewards consistent sampled answers~\cite{zuo2025ttrl,ttrv}.
Closest to InCue is the test-time optimization of latents, which updates only a learnable offset on the visual tokens, or a few extra latent tokens, toward an objective: attention inside a user-given region~\cite{wu2024controlmllm,wu2026controlmllmpp}, the answer logit of sub-queries~\cite{yang2025lra}, the activations of a base model~\cite{xing2026activationreplay}, or reasoning confidence~\cite{liu2026dmlr,zhang2026visuallatents}.
InCue also optimizes only an offset on the visual tokens, but its target is spatial, built from the model's own attention and entropy gradient, and rebuilt at every step. It therefore needs no externally provided region, reference model, or sampling.

\section{Method}

\subsection{Overview}
Fig.~\ref{fig:main_incue} depicts the overall framework of InCue. With the MLLM frozen,
InCue optimizes only a learnable additive offset on the cached visual tokens. It first
reads out two internal cues from the model, the attention cue and the sensitivity cue
(Sec.~\ref{sec:prelim}). Spatial Cue Coupling (SCC, Sec.~\ref{sec:scc}) couples them into
a joint response, and Regional Cue Refocusing (RCR, Sec.~\ref{sec:rcr}) refocuses this
response on its strongest region to form the spatial target. Test-time optimization
(Sec.~\ref{sec:tto}) then updates the offset to steer the attention toward the target,
refreshing both cues and the target at every step, and the answer is read out after the
last step.
\subsection{Preliminaries}
\label{sec:prelim}

InCue operates on a frozen MLLM consisting of a vision encoder, a projector, and an LLM
decoder, with parameters $\theta$. Given an image $I$ and a text prompt $x$ containing
the question and the candidate options, the vision encoder and the projector, jointly
denoted as $\mathcal{E}$, encode the image into $N$ visual tokens
\begin{equation}
U_0=[u_1,\dots,u_N]^\top=\mathcal{E}(I)\in\mathbb{R}^{N\times D},
\end{equation}
where $U_0$ is the visual input of the LLM decoder, each token $u_i$ corresponds to a
local patch of the image, and $D$ is the hidden
dimension of the LLM. These tokens are concatenated with the text tokens and processed
by the decoder. The image is encoded only once: $U_0$ is cached, and everything that
follows operates on these tokens rather than on the image.

With $\theta$ and the prompt fixed, everything the decoder computes depends only on the
visual tokens it receives. A learnable additive offset on the cached tokens,
$U=U_0+\Delta U$, is therefore enough to change how the model uses its already-encoded
visual information, without new visual inputs or any change to the model. To decide
where to steer the attention, InCue reads out two internal cues from the model at the
current $U$.

The attention cue describes where the model currently looks. Let $\mathcal{Q}$ be the
set of text positions after the image, i.e., those of the question, the options, and
the answer prompt, and let $\alpha^{(\ell,m)}_{q,i}(U)$ denote the attention weight that
position $q\in\mathcal{Q}$ assigns to visual token $i$ in the $m$-th head of the
$\ell$-th decoder layer. The cue pools this text-to-visual attention over a fixed set
$\Lambda$ of read-out layers (e.g., layers 15--26 in Qwen2.5-VL-7B), all $M$ heads, and
the text positions $\mathcal{Q}$,
\begin{equation}
A=\mathrm{Norm}\Big(\sum_{\ell\in\Lambda}\sum_{m=1}^{M}\sum_{q\in\mathcal{Q}}
\alpha^{(\ell,m)}_{q,\cdot}(U)\Big),
\end{equation}
where $\mathrm{Norm}(v)=v/\sum_j v_j$ rescales a nonnegative vector to sum to one.

The sensitivity cue instead asks where the model's uncertainty reacts. At the output, a
softmax over the option logits at the first decoding step gives the predictive
distribution $p(U)$ over the candidate answers in $\mathcal{Y}$, whose Shannon entropy
\begin{equation}
H(p)=-\sum_{y\in\mathcal{Y}}p_y\log p_y
\end{equation}
quantifies the model's uncertainty among them; we write it as $H(U)$. The cue
backpropagates this entropy to each visual token, as in EGG~\cite{gropl2026egg}, and
keeps the norm of the gradient,
\begin{equation}
g_i=\frac{\partial H(U)}{\partial u_i}, \
S=\mathrm{Norm}\big([\|g_1\|_2,\dots,\|g_N\|_2]^\top\big),
\end{equation}
so $S_i$ is large when a small perturbation of $u_i$ changes $H$ strongly, in either
direction.

\subsection{Spatial Cue Coupling}
\label{sec:scc}
Steering the attention with the offset $\Delta U$ requires a spatial target over the
visual tokens. SCC builds its basis, the joint response $J$, from the two cues.
They are complementary: attention measures how visual information is currently
routed, while sensitivity measures how strongly changes to it would affect the
predictive uncertainty. Although they stem from the same computation, they are not
interchangeable: a highly attended token may barely move the uncertainty because its
information is redundant, whereas a sensitive token may receive little attention in the
read-out layers. SCC therefore couples them through the element-wise product,
\begin{equation}
J=\mathrm{Norm}(A\odot S),
\end{equation}
which keeps a position prominent only when both cues support it
(Fig.~\ref{fig:main_incue}). To first order, $A_iS_i$ is, up to a constant, the largest
entropy change that perturbing $u_i$ can cause within a budget proportional to its
attention. Put differently, $J$ is the current attention reweighted by sensitivity, so
it stays within reach of the attention, unlike the entropy gradient alone
(Fig.~\ref{fig:steer}d).
\subsection{Regional Cue Refocusing}
\label{sec:rcr}
The joint response scores each visual position independently, so its mass can still
scatter across the image, whereas the evidence for a question typically occupies a
contiguous region. RCR therefore refocuses the joint response on the region where it is
strongest. Following the window-based selection of prior
work~\cite{zhang2025vicrop,gropl2026egg}, it slides a fixed-size window over the image
to form a candidate set $\mathcal{W}$ and scores each window by the joint response it
covers,
\begin{equation}
\phi(W)=\sum_{i=1}^{N}\omega_i(W)\,J_i,\qquad
W^\star=\arg\max_{W\in\mathcal{W}}\phi(W),
\end{equation}
where $\omega_i(W)\in[0,1]$ is the fraction of patch $i$ covered by window $W$.

RCR then amplifies the joint response inside the selected window to form the spatial
target,
\begin{equation}
T=\mathrm{Norm}\big(\big[J_i\,\rho^{\,\mathbb{I}[i\in W^\star]}\big]_{i=1}^{N}\big),
\end{equation}
where $i\in W^\star$ means that the center of patch $i$ falls inside the window and
$\rho>1$ is a constant. Because $T$ is renormalized, $\rho$ acts exactly on the odds of
the window: the ratio of target mass inside $W^\star$ to that outside is $\rho$ times
its value under $J$, while the relative weights within each side stay unchanged. The
fine-grained profile of $J$ is thus preserved, and $\rho=1$ recovers $T=J$. Unlike
re-presentation, the window only shapes the target: nothing is cropped or re-encoded, no
visual tokens are added, and the region keeps its surrounding context.

\subsection{Test-Time Optimization}
\label{sec:tto}
InCue steers the attention toward the target by updating the offset $\Delta U$ on the visual tokens.
At step
$k$, the visual representations are $U^{(k)}=U_0+\Delta U^{(k)}$, and the cues and the
target are rebuilt from them following Secs.~\ref{sec:prelim}--\ref{sec:rcr}, written
$A^{(k)}$, $S^{(k)}$, $J^{(k)}$, and $T^{(k)}$. The objective measures how much
attention falls on the target,
\begin{equation}
\mathcal{L}^{(k)}=-\log \big\|A^{(k)}\odot \stopgrad[T^{(k)}]\big\|_\gamma,
\end{equation}
where $\stopgrad$ denotes the stop-gradient operator, $\gamma\ge1$ controls how strongly the largest co-activations dominate, and the
log keeps the gradient strong while the overlap is still small.
Since $T^{(k)}$ is built from $S^{(k)}$, which already contains the gradient of $H$, differentiating through the target would require second-order derivatives. We therefore treat $T^{(k)}$ as a constant within each update ($\stopgrad$), so the gradient reaches the offset only through the attention:
\begin{equation}
\Delta U^{(k+1)}=\Delta U^{(k)}-\eta\,\nabla_{\Delta U}\mathcal{L}^{(k)},
\end{equation}
where $\eta$ is the step size.

For intuition, take $\gamma=1$ and $\rho=1$: minimizing $\mathcal{L}^{(k)}$ then amounts to maximizing
\begin{equation}
\sum_i A_i(U)\,\stopgrad\!\left[\frac{A_i^{(k)}S_i^{(k)}}{\sum_j A_j^{(k)}S_j^{(k)}}\right],
\end{equation}
where $A_i(U)$ is the attention computed from the tokens being updated, and the bracket is the target, frozen at step $k$. Each update thus moves the attention toward the positions that were both attended and sensitive at step $k$. A larger $\gamma$ favors the strongest of them. Rebuilding the window and the target at every step matters because each update changes what the model attends to, predicts, and is
sensitive to; a target built once at the start would soon describe a state that no
longer exists. After $K$ (usually 6) steps, the answer is read out as
$\hat{y}=\arg\max_y p_y(U^{(K)})$.

\section{Experiments}
\label{sec:experiment}

\begin{table}[!t]
\centering
\setlength{\heavyrulewidth}{1.1pt}\setlength{\lightrulewidth}{0.6pt}\setlength{\cmidrulewidth}{0.4pt}
\caption{\textbf{Comparison with test-time methods under matched visual-token input budgets.}
For a fair comparison, all methods in a block use the same base model and the same
visual tokens, with the budget in the left column.
We run AIF~\cite{liu2026aif} and TTRV~\cite{ttrv} (both CVPR'26) with their official code. Every method is shown at its best setting and step. (Supp. fixes one setting per method). TTRV runs out of memory at
native resolution. Subscripts give the gain over the base model, and Avg.\ is the mean over the three benchmarks. Ours is shaded in gray.}
\label{tab:main}
\renewcommand{\arraystretch}{1.1}
\setlength{\tabcolsep}{2.6pt}
\begin{adjustbox}{max width=\columnwidth}
\begin{tabular}{@{}c l c c c c@{}}
\toprule
Tokens & Method & V$^*$ & HR-4K & MME-RW & Avg. \\
\midrule
\multirow{3}{*}{\shortstack{$\le$16K\\(native)}} & Qwen2.5-VL-7B & 79.1 & 71.8 & 52.7 & 67.8 \\
 & +\,AIF~\cite{liu2026aif} & 81.2 & 72.5 & 53.4 & 69.0 \\
 & \cellcolor{oursbg}+\,\textbf{InCue} & \cellcolor{oursbg}\textbf{82.7}\gain{+3.7} & \cellcolor{oursbg}\textbf{73.1}\gain{+1.4} & \cellcolor{oursbg}\textbf{55.6}\gain{+3.0} & \cellcolor{oursbg}\textbf{70.5}\gain{+2.7}\\
\midrule
\multirow{4}{*}{\shortstack{2048}} & Qwen2.5-VL-7B & 68.6 & 64.0 & 48.6 & 60.4 \\
 & +\,AIF~\cite{liu2026aif} & 71.7 & 66.1 & 48.8 & 62.2 \\
 & +\,TTRV~\cite{ttrv} & 71.0 & 66.1 & 48.2 & 61.8 \\
 & \cellcolor{oursbg}+\,\textbf{InCue} & \cellcolor{oursbg}\textbf{73.8}\gain{+5.2} & \cellcolor{oursbg}\textbf{66.3}\gain{+2.3} & \cellcolor{oursbg}\textbf{51.8}\gain{+3.3} & \cellcolor{oursbg}\textbf{64.0}\gain{+3.6} \\
\midrule
\multirow{4}{*}{\shortstack{1024}} & Qwen2.5-VL-7B & 70.7 & 61.9 & 46.6 & 59.7 \\
 & +\,AIF~\cite{liu2026aif} & 74.3 & 64.0 & 49.2 & 62.5 \\
 & +\,TTRV~\cite{ttrv} & 73.3 & 62.4 & 46.9 & 60.9 \\
 & \cellcolor{oursbg}+\,\textbf{InCue} & \cellcolor{oursbg}\textbf{75.9}\gain{+5.2} & \cellcolor{oursbg}\textbf{65.6}\gain{+3.8} & \cellcolor{oursbg}\textbf{51.5}\gain{+5.0} & \cellcolor{oursbg}\textbf{64.4}\gain{+4.7} \\
\midrule
\multirow{6}{*}{\shortstack{576}} & Qwen2.5-VL-7B & 61.3 & 57.3 & 42.6 & 53.7 \\
 & +\,AIF~\cite{liu2026aif} & 62.8 & 58.4 & 44.2 & 55.1 \\
 & +\,TTRV~\cite{ttrv} & 63.5 & 58.6 & 42.9 & 55.0 \\
 & \cellcolor{oursbg}+\,\textbf{InCue} & \cellcolor{oursbg}\textbf{66.5}\gain{+5.2} & \cellcolor{oursbg}\textbf{60.0}\gain{+2.8} & \cellcolor{oursbg}\textbf{49.1}\gain{+6.5} & \cellcolor{oursbg}\textbf{58.5}\gain{+4.8} \\
\cmidrule(l){2-6}
 & Qwen2-VL-7B & 56.0 & 59.9 & 46.0 & 54.0 \\
 & \cellcolor{oursbg}+\,\textbf{InCue} & \cellcolor{oursbg}\textbf{61.8}\gain{+5.8} & \cellcolor{oursbg}\textbf{60.1}\gain{+0.3} & \cellcolor{oursbg}\textbf{48.0}\gain{+2.0} & \cellcolor{oursbg}\textbf{56.6}\gain{+2.7} \\
\midrule
\multirow{3}{*}{\shortstack{$\le$3328\\($\le$12 tiles)}} & InternVL3-8B & 70.7 & 71.3 & 55.1 & 65.7 \\
 & +\,AIF~\cite{liu2026aif} & 74.9 & 73.4 & 57.6 & 68.6 \\
 & \cellcolor{oursbg}+\,\textbf{InCue} & \cellcolor{oursbg}\textbf{79.6}\gain{+8.9} & \cellcolor{oursbg}72.4\gain{+1.1} & \cellcolor{oursbg}\textbf{58.8}\gain{+3.7} & \cellcolor{oursbg}\textbf{70.3}\gain{+4.6} \\
\midrule
\multirow{3}{*}{\shortstack{$\le$2304\\($\le$8 tiles)}} & InternVL3-8B & 69.1 & 62.3 & 55.8 & 62.4 \\
 & +\,AIF~\cite{liu2026aif} & 72.8 & 63.3 & 57.3 & 64.4 \\
 & \cellcolor{oursbg}+\,\textbf{InCue} & \cellcolor{oursbg}\textbf{77.5}\gain{+8.4} & \cellcolor{oursbg}\textbf{64.8}\gain{+2.5} & \cellcolor{oursbg}\textbf{58.9}\gain{+3.1} & \cellcolor{oursbg}\textbf{67.0}\gain{+4.6} \\
\midrule
\multirow{7}{*}{\shortstack{256\\(1 tile)}} & InternVL3-8B & 59.2 & 57.3 & 46.6 & 54.3 \\
 & +\,AIF~\cite{liu2026aif} & 61.8 & 57.9 & 46.5 & 55.4 \\
 & +\,TTRV~\cite{ttrv} & 59.2 & 57.3 & 46.9 & 54.4 \\
 & \cellcolor{oursbg}+\,\textbf{InCue} & \cellcolor{oursbg}\textbf{62.8}\gain{+3.7} & \cellcolor{oursbg}\textbf{60.6}\gain{+3.4} & \cellcolor{oursbg}\textbf{48.4}\gain{+1.8} & \cellcolor{oursbg}\textbf{57.3}\gain{+3.0} \\
\cmidrule(l){2-6}
 & InternVL3.5-8B & 51.3 & 51.6 & 42.8 & 48.6 \\
 & +\,AIF~\cite{liu2026aif} & 51.8 & 50.6 & 43.1 & 48.5 \\
 & \cellcolor{oursbg}+\,\textbf{InCue} & \cellcolor{oursbg}\textbf{55.0}\gain{+3.7} & \cellcolor{oursbg}\textbf{52.8}\gain{+1.1} & \cellcolor{oursbg}\textbf{43.5}\gain{+0.7} & \cellcolor{oursbg}\textbf{50.4}\gain{+1.8}\\
\bottomrule
\end{tabular}
\end{adjustbox}
\end{table}

\begin{figure*}[!t]
  \centering
  \includegraphics[width=\textwidth]{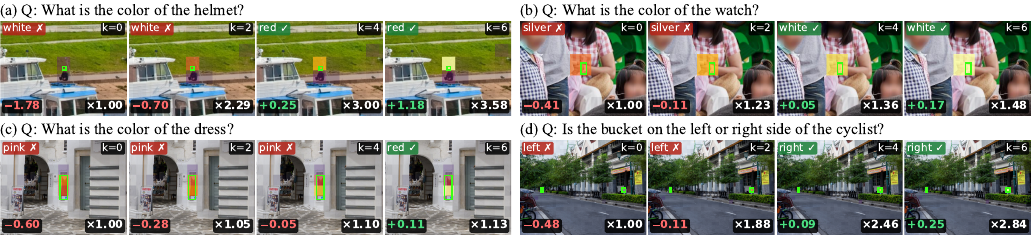}
  \caption{\textbf{InCue corrects the answer step by step.}
 Four V$^*$ questions that Qwen2.5-VL-7B answers wrongly, shown at steps 0, 2, 4 and 6 of InCue.
Heat maps show attention over visual tokens, green boxes the ground-truth targets, and
top-left labels the predictions (green correct, red wrong). Bottom-left numbers give the
correct-answer margin (correct minus strongest wrong option logit), and bottom-right numbers
the attention in the boxes relative to step 0. Attention on the target and the margin rise
together until the answer flips.}
  \label{fig:rescue_steps}
\end{figure*}

\begin{table}[!t]
\centering
\setlength{\heavyrulewidth}{1.1pt}\setlength{\lightrulewidth}{0.6pt}\setlength{\cmidrulewidth}{0.4pt}
\caption{\textbf{Comparison with crop-based methods} (Qwen2.5-VL-7B).
Following ViCrop's official setting, each method appends crops of the original image, selected by its
region cue, to the 576-token input of Tab.~\ref{tab:main}. ``Tuned'' reports the strongest attention
layer and crop size per benchmark, as no complete official setting exists for this model (EGG and InCue
crop likewise use their strongest crop rule). Top: \textbf{InCue crop} selects the crop with our joint
response $J$, then applies InCue ($K{=}0$: no optimization). Middle: InCue on each prior method's crop.
Bottom: $K{=}0$ accuracy per region cue (row) and crop rule (column), averaged over the three benchmarks;
boxes follow ViCrop's rule with the given base size. Subscripts: gain over the strongest prior method per
column (top) or over that method alone (middle).}
\label{tab:crop}
\renewcommand{\arraystretch}{1.1}
\setlength{\tabcolsep}{3pt}
\begin{adjustbox}{max width=\columnwidth}
\begin{tabular}{@{}l c c c c c@{}}
\toprule
Crop-based methods & Venue & V$^*$ & HR-4K & MME-RW & Avg. \\
\midrule
Qwen2.5-VL-7B~\cite{bai2025qwen25vl} & arXiv'25 & 61.3 & 57.3 & 42.6 & 53.7 \\
+\,ViCrop~\cite{zhang2025vicrop} & ICLR'25 & 62.8 & 57.4 & 43.9 & 54.7 \\
+\,ViCrop~\cite{zhang2025vicrop} (tuned) & ICLR'25 & 75.9 & 72.1 & 52.5 & 66.8 \\
+\,HiDe~\cite{liu2026hide} & ICML'26 & 69.1 & 65.3 & 50.2 & 61.5 \\
+\,EGG~\cite{gropl2026egg} & ECCV'26 & 72.8 & 70.3 & 52.6 & 65.2 \\
\rowcolor{oursbg}+\,\textbf{InCue crop} ($K{=}0$) & Submission & 76.4\gain{+0.5} & 73.4\gain{+1.3} & 53.3\gain{+0.7} & 67.7\gain{+0.8} \\
\rowcolor{oursbg}+\,\textbf{InCue crop} & Submission & \textbf{77.5}\gain{+1.6} & \textbf{73.6}\gain{+1.5} & \textbf{53.8}\gain{+1.2} & \textbf{68.3}\gain{+1.5} \\
\midrule[\heavyrulewidth]
+\,ViCrop + \textbf{InCue} & & 64.9\gain{+2.1} & 59.5\gain{+2.1} & 46.0\gain{+2.1} & 56.8\gain{+2.1} \\
+\,ViCrop (tuned) + \textbf{InCue} & & 76.4\gain{+0.5} & 74.3\gain{+2.1} & 53.1\gain{+0.6} & 67.9\gain{+1.1} \\
+\,HiDe + \textbf{InCue} & & 74.3\gain{+5.2} & 67.6\gain{+2.4} & 51.7\gain{+1.5} & 64.6\gain{+3.0} \\
+\,EGG + \textbf{InCue} & & 74.3\gain{+1.6} & 70.8\gain{+0.5} & 53.3\gain{+0.7} & 66.1\gain{+0.9} \\
\midrule[\heavyrulewidth]
\multirow{2}{*}{Region cue} & \multicolumn{2}{c}{Window (side)} & \multicolumn{3}{c}{Box (px)} \\
\cmidrule(lr){2-3}\cmidrule(l){4-6}
 & 1/8 & 1/2 & 336 & 448 & 672 \\
\midrule
ViCrop & 54.6 & 58.4 & 55.7 & 55.9 & 58.3 \\
ViCrop (tuned) & 62.8 & 64.7 & 62.4 & 65.1 & 64.6 \\
HiDe & 57.0 & 62.2 & 58.0 & 60.2 & 62.2 \\
EGG & 60.5 & 63.0 & 61.2 & 62.8 & 63.9 \\
\rowcolor{oursbg}\textbf{InCue} & \textbf{63.3} & \textbf{65.4} & \textbf{62.6} & \textbf{65.2} & \textbf{66.0} \\
\bottomrule
\end{tabular}
\end{adjustbox}
\end{table}

\paragraph{Benchmarks.}
We evaluate on three fine-grained perception benchmarks whose questions hinge on small regions of high-resolution images, V$^*$~\cite{wu2024vstar}, HR-Bench~4K~\cite{wang2024hrbench} and the perception split of MME-RealWorld-Lite~\cite{zhang2024mmerealworld}, and on eight general benchmarks: MMVP~\cite{tong2024mmvp}, POPE~\cite{li2023pope}, CV-Bench~\cite{tong2024cambrian}, RealWorldQA~\cite{xai2024realworldqa}, MMStar~\cite{chen2024mmstar}, MathVista~\cite{lu2024mathvista}, SEED-Bench~\cite{li2023seed}, and COCO-Exist, 2{,}016 binary POPE-style object-existence questions on COCO images~\cite{lin2014coco}. All are multiple-choice, and we report accuracy; subset sizes and construction details are in the Supp.

\paragraph{Implementation details.}
We evaluate four base models from two families: Qwen2-VL-7B~\cite{wang2024qwen2vl}, Qwen2.5-VL-7B~\cite{bai2025qwen25vl}, InternVL3-8B~\cite{zhu2025internvl3} and InternVL3.5-8B~\cite{wang2025internvl35}. Test-time and crop-based baselines use their official code, the latter ported to each base model. InCue optimizes for at most $K$ steps; its hyper-parameters (read-out layers $\Lambda$, step size $\eta$, odds $\rho$, norm $\gamma$), their search ranges, and the computational cost (latency, peak memory) of all methods are given in the Supp. Experiments were run on 4$\times$ A100-80GB GPUs.

\paragraph{Evaluation protocol.}
All methods share one protocol. Each comparison fixes the visual-token budget, downsampling rather than discarding larger images. For a fair comparison, Tab.~\ref{tab:main} reports every method at the best setting and step we ran; with one fixed setting per method instead (InCue: one step size per base model and $K{=}6$ steps everywhere), InCue still achieves the best average accuracy in every setting, with a mean gain of 2.2\% (Supp.). All methods use the official chat template and the same readout, the highest-logit option letter, which reduces readout-induced gaps: it avoids missed or misread answers and is robust to format changes from test-time training, so accuracy differences mainly reflect the model's choice. The protocol also avoids issues in some prior evaluations: a malformed chat template that leaves many responses empty and lowers the baseline well below its official accuracy, a parser that takes the last option letter and can misread free-form answers mentioning several options, and a main-table result that appears to use more adaptation samples than stated.

\subsection{Comparison with State-of-the-art Methods}
\label{sec:exp_main}
We compare InCue with recent test-time methods in Tab.~\ref{tab:main}. InCue improves every
base model on every benchmark at every budget and achieves the best average accuracy in all
settings. The gains are typically largest on V$^*$ and reach 8.9\% on InternVL3-8B. On
Qwen2.5-VL-7B, the average gain grows as the budget shrinks, from 2.7\% at native resolution
to 4.7\% at 1024 and 4.8\% at 576 tokens. At native resolution, InCue still improves V$^*$ by 3.7\%
and MME-RW by 3.0\%, so the model leaves part of its visual evidence unused even with the
full image. As a reference upper bound, we also run InCue on V$^*$ with the ground-truth box
as the target, and its internal cues alone already recover most of this oracle's gain (see
Supp.).
\begin{table*}[!t]
\centering
\setlength{\heavyrulewidth}{1.1pt}\setlength{\lightrulewidth}{0.6pt}\setlength{\cmidrulewidth}{0.4pt}
\caption{\textbf{Results on general multimodal benchmarks.}
All benchmarks use at most 576 visual tokens. Subscripts give the gain over the base model, and Avg.\ is the mean over the eight benchmarks. InCue improves all eight, although they depend less on small details.}
\label{tab:general}
\setlength{\tabcolsep}{4pt}
\begin{adjustbox}{max width=\textwidth}
\begin{tabular}{@{}l ccccccccc@{}}
\toprule
Method & MMVP & POPE & CV-Bench & RealWorldQA & MMStar & MathVista & SEED-Bench & COCO-Exist & Avg. \\
\midrule
Qwen2.5-VL-7B & 54.0 & 84.9 & 80.6 & 71.4 & 62.0 & 78.7 & 75.7 & 87.9 & 74.4 \\
\rowcolor{oursbg}\,+ \textbf{InCue (Ours)} & 56.7\gain{+2.7} &  87.0\gain{+2.2} & 81.4\gain{+0.8} & 73.2\gain{+1.8} & 62.3\gain{+0.3}& 79.6\gain{+0.9} & 76.5\gain{+0.8} & 88.9\gain{+1.0} & 75.7\gain{+1.3} \\
\bottomrule
\end{tabular}
\end{adjustbox}
\end{table*}

\begin{table}[!t]
\centering
\setlength{\heavyrulewidth}{1.1pt}\setlength{\lightrulewidth}{0.6pt}\setlength{\cmidrulewidth}{0.4pt}
\setlength{\aboverulesep}{0pt}\setlength{\belowrulesep}{0pt}
\caption{\textbf{Gains concentrate on hard samples.}
Hard samples are the third of each full set with the smallest confidence gap of the base model (top-1 minus top-2 option logit), selected without labels by a rule fixed before evaluation,
separately per number of options. $\Delta$ is the gain over the base model on the same
questions (all or hard), and AIF and TTRV use their settings from Tab.~\ref{tab:main}.}
\label{tab:hard}
\footnotesize
\setlength{\tabcolsep}{3pt}
\renewcommand{\arraystretch}{1.2}
\begin{adjustbox}{width=\columnwidth}
\begin{tabular}{@{}c|c l c c c c c c@{}}
\toprule
\multicolumn{1}{@{}c}{\multirow{2}{*}{Model}} & \multirow{2}{*}{Tokens} & \multirow{2}{*}{Bench.} & \multicolumn{2}{c}{Base acc.} & \multicolumn{2}{c}{$\Delta$ (hard)} & \multicolumn{2}{c}{InCue $\Delta$} \\
\cmidrule(lr){4-5}\cmidrule(lr){6-7}\cmidrule(l){8-9}
\multicolumn{1}{@{}c}{} & & & all & hard & AIF & TTRV & all & hard \\
\midrule
\multirow{3}{*}{Qwen2.5-VL} & 2048 & V$^*$ & 68.6 & 43.8 & +9.4 & +6.3 & \textcolor{gaingreen}{+5.2} & \textbf{\textcolor{gaingreen}{+12.5}} \\
 & 576 & V$^*$ & 61.3 & 42.2 & +4.7 & +6.3 & \textcolor{gaingreen}{+5.2} & \textbf{\textcolor{gaingreen}{+12.5}} \\
 & 576 & MME-RW & 42.6 & 30.9 & +3.7 & +0.7 & \textcolor{gaingreen}{+6.5} & \textbf{\textcolor{gaingreen}{+8.9}} \\
\midrule
\multirow{2}{*}{InternVL3} & $\le$3328 & V$^*$ & 70.7 & 34.4 & +12.5 & -- & \textcolor{gaingreen}{+8.9} & \textbf{\textcolor{gaingreen}{+23.4}} \\
 & $\le$2304 & V$^*$ & 69.1 & 29.7 & +9.4 & -- & \textcolor{gaingreen}{+8.4} & \textbf{\textcolor{gaingreen}{+21.9}} \\
\midrule
InternVL3.5 & 256 & V$^*$ & 51.3 & 35.9 & +1.6 & -- & \textcolor{gaingreen}{+3.7} & \textbf{\textcolor{gaingreen}{+9.4}} \\
\bottomrule
\end{tabular}
\end{adjustbox}
\end{table}

\begin{figure}[!t]
    \centering
    \includegraphics[width=\linewidth]{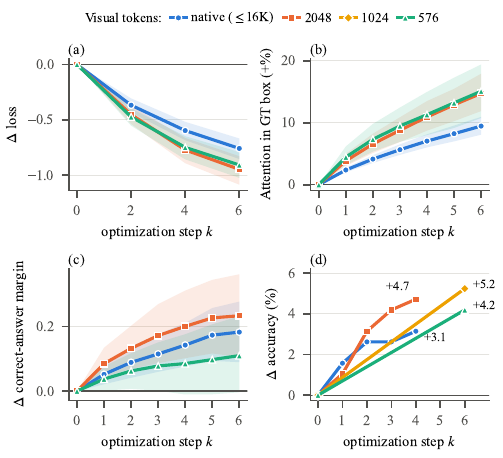}
    \caption{\textbf{Attention, margin, and accuracy rise together during optimization} (V$^*$,
Qwen2.5-VL-7B). (a)--(c) Averages over all questions during $K{=}6$ InCue steps at three budgets, with 95\% confidence intervals shaded. (a) Change of $\mathcal{L}$. (b) relative increase of attention in the ground-truth boxes, used only for
evaluation; (c) Change of the correct-answer margin (correct minus strongest wrong option
logit). (d) Accuracy gain over the base model.}
    \label{fig:vstar_curves}
\end{figure}

\subsection{Comparison with Crop-Based Methods}
\label{sec:exp_crop}

We compare InCue with crop-based methods in Tab.~\ref{tab:crop}. Crop-based methods add an
enlarged region of the original image to the input, whereas InCue leaves the input unchanged
and only changes how the model uses it, so the two are compatible. For a fair comparison under
the same input, we build InCue crop, which appends a crop of the original image selected by our joint response $J$, as these methods do, and then applies InCue. This is the only setting in which InCue crops: everywhere else, the window selected from $J$ only reweights the spatial target (Sec.~3.4), and the input is left unchanged. Every row of Tab.~\ref{tab:crop} thus pays for the extra crop with
additional visual tokens, whereas InCue in Tab.~\ref{tab:main} adds none. InCue crop outperforms all crop-based methods on all three benchmarks (top), averaging 68.3\%, higher than ViCrop tuned to its strongest attention layer and crop size (66.8\%)
and EGG (65.2\%). Applied to each prior method's own crop (middle), InCue also improves every method on every
benchmark, by up to 5.2\% (HiDe on V$^*$) and by 2.1\% even for the tuned ViCrop on HR-4K.
The gain persists at the native budget of Tab.~\ref{tab:main} (Supp.): on V$^*$, InCue raises
ViCrop from 78.0\% to 80.6\%, HiDe from 91.6\% to 92.7\%, and the crop selected by $J$ from
84.3\% to 86.4\%. Cropping and InCue thus act at different stages, on the input and on its use, and their gains combine even on top of a tuned crop.

Since crop size strongly affects accuracy, the bottom of Tab.~\ref{tab:crop} compares the
region cues under the same crop rule, with everything else fixed and no optimization. Under all five crop rules, cropping with our joint response $J$ gives the
highest average accuracy and surpasses the tuned ViCrop, which shows the effectiveness of coupling
the internal cues. In contrast, ViCrop with its official setting gains only 1.0\% and becomes
competitive only after its attention layer and crop size are tuned for each benchmark. Full
per-benchmark results are in the Supp.

\subsection{Results on General Benchmarks}
\label{sec:exp_general}
Beyond high-resolution questions, we test InCue's generalization on eight general
benchmarks (Tab.~\ref{tab:general}). InCue improves the base model on all of them, by
1.3\% on average. The largest gains appear on MMVP (+2.7\%) and POPE (+2.2\%), both of
which require the model to attend to the actual evidence in the image rather than to
obtain more detail, suggesting that steering attention toward evidence the model
already has also helps mitigate hallucination.

\begin{figure*}[!t]
  \centering
  \includegraphics[width=\textwidth]{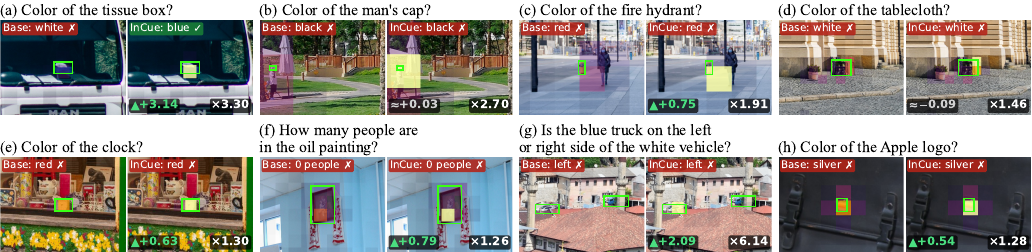}
\caption{\textbf{When does stronger attention fail?} Eight V$^*$ examples on Qwen2.5-VL-7B
(576/2048 tokens or native) before (left) and after (right) six InCue steps. Heat maps show
attention, green boxes the ground-truth targets, and top-left labels the predictions (green correct,
red wrong). Bottom-left: change in the correct-answer margin (correct minus strongest wrong option
logit, negative while the answer is wrong, so a rise moves toward the correct answer); bottom-right:
attention in the box relative to step 0.
\textcolor[RGB]{0,128,64}{$\blacktriangle$}: rise; $\approx$: within $\pm$0.1.}

  \label{fig:failure}
\end{figure*}

\begin{table}[!t]
\centering
\setlength{\heavyrulewidth}{1.1pt}\setlength{\lightrulewidth}{0.6pt}\setlength{\cmidrulewidth}{0.4pt}
\setlength{\aboverulesep}{0pt}\setlength{\belowrulesep}{0pt}
\caption{\textbf{Ablation of InCue's components.} Top: targets from different cues (Qwen2.5-VL-7B,
2048 tokens). Bottom: adding RCR and per-step cue refresh (Qwen2.5-VL-7B on MME-RW, InternVL3.5-8B
with 1 tile on V$^*$). Each block shares all other settings; we report the best accuracy within
$K{=}6$ steps ($K{=}12$ for MME-RW at the top). Subscripts: gain over the base model.}
\label{tab:ablation}
\footnotesize
\setlength{\tabcolsep}{3pt}
\renewcommand{\arraystretch}{1.2}
\begin{tabularx}{\columnwidth}{@{}c|>{\raggedright\arraybackslash}X c c@{}}
\toprule
\multicolumn{1}{@{}c}{} & Variant & MME-RW & V$^*$ \\
\midrule
\multirow{4}{*}{\shortstack{\textit{Cue}\\\textit{design}}} & Base (Qwen2.5-VL-7B) & 48.6\phantom{\gain{+0.0}} & 68.6\phantom{\gain{+0.0}} \\
 & Attention cue only ($A$) & 50.3\gain{+1.7} & 70.7\gain{+2.1} \\
 & Sensitivity cue only ($S$) & 51.6\gain{+3.1} & 72.8\gain{+4.2} \\
 & \cellcolor{oursbg}Joint cue $J$ & \cellcolor{oursbg}\textbf{51.6}\gain{+3.1} & \cellcolor{oursbg}\textbf{73.3}\gain{+4.7} \\
\midrule
\multirow{4}{*}{\shortstack{\textit{RCR \&}\\\textit{refresh}}} & Base (Qwen2.5-VL / InternVL3.5-8B) & 48.6\phantom{\gain{+0.0}} & 51.3\phantom{\gain{+0.0}} \\
 & $J$, fixed target ($\rho{=}1$) & 49.8\gain{+1.2} & 53.9\gain{+2.6} \\
 & + RCR ($\rho{=}4$) & 50.4\gain{+1.8} & 54.5\gain{+3.1} \\
 & \cellcolor{oursbg}+ cue refresh & \cellcolor{oursbg}\textbf{50.6}\gain{+2.0} & \cellcolor{oursbg}\textbf{55.0}\gain{+3.7} \\
\bottomrule
\end{tabularx}
\end{table}

\subsection{Hard-Sample Analysis}
\label{sec:exp_hard}
Since InCue is driven by the model's predictive uncertainty, we test how much it helps on
questions where the base model hesitates. Without using labels, we take the third of each set with the smallest confidence gap of the base model as hard samples. As Tab.~\ref{tab:hard} shows, in every setting InCue's gain on hard samples is considerably larger than on the full set, typically two to
three times as large and up to 23.4\% on InternVL3-8B, and exceeds the gains of AIF and TTRV. This is because on hard samples the top two options are separated by only a small logit gap, so the moderate change brought by InCue
is enough to flip the answer. Fig.~\ref{fig:rescue_steps} illustrates this
process: as InCue raises the attention on the ground-truth target box step by step, the correct-answer margin grows until the answer flips.

\subsection{Optimization Dynamics}
\label{sec:exp_dynamics}
We track the optimization of InCue on V$^*$ in Fig.~\ref{fig:vstar_curves}. As the loss
decreases, the attention inside the target boxes keeps rising, by 9.5\%, 14.7\%, and 15.1\%
after six steps at native resolution and at 2048 and 576 tokens. The correct-answer margin
grows monotonically with it, and accuracy improves by 3.1--5.2\% within six steps.
\textbf{Notably,
the loss of InCue uses neither labels nor target boxes, yet the attention moves steadily
onto the annotated targets, showing that the spatial target built from internal cues
indeed points to the evidence needed to answer the question.}
The smaller the budget, the
larger the increase in attention, in line with the larger gains at smaller budgets in
Tab.~\ref{tab:main}.

\subsection{Failure Analysis}
\label{sec:exp_failure}
Stronger attention does not always lead to the correct answer, and
Fig.~\ref{fig:failure} contrasts a corrected case (a) with typical failures. First, when the
target is extremely small, the visual tokens have already lost the needed detail (b), so the
margin barely changes even when the attention concentrates on the target. This mainly happens
at low budgets such as 576 tokens. Second, language priors can override the visual evidence,
\eg, the model assumes that fire hydrants are red and tablecloths are white (c, d). Third,
when the model starts from a confident wrong answer, the correct option trails too far
behind: its margin rises, by up to 2.09 in (g), but does not flip within six steps (e--h).

\subsection{Ablation Study}
\label{sec:exp_ablation}
The top of Tab.~\ref{tab:ablation} compares target sources: the joint response is best on both
benchmarks, whereas neither cue alone matches it on both. Unlike a KL loss, which fits the attention
to the whole target (to $S$ in Fig.~\ref{fig:steer}d), our overlap only pulls attention to the
target's peaks, so, as in region selection, a cue only needs to peak at the right place. The bottom
starts from a fixed $J$ computed at step 0 and adds RCR and cue refresh, which recomputes the cues
and target at every step, each improving both benchmarks.

\section{Conclusion}
We presented InCue, which couples the model’s internal cues into a spatial target and
optimizes only an offset on visual tokens at test time, steering attention
toward evidence the model has already seen.
The cues build on the observation that positions where both attention and the entropy
gradient are high reveal the spatial target. Internal cues easily give a coarse location, but
using the evidence requires turning them into a target that attention can follow, which we
achieve by coupling the cues and refocusing them on a region. Our approach makes better
use of evidence under the same budget, complements cropping, and gains most on hard samples in comparison to SOTA. Relying on internal cues, InCue needs no
additional input but cannot introduce new evidence, so confident errors remain hard to
correct.

{
    \small
    \bibliographystyle{ieeenat_fullname}
    \bibliography{main}
}
\clearpage
\renewcommand{\thetable}{S\arabic{table}}
\renewcommand{\thefigure}{S\arabic{figure}}
\renewcommand{\thealgorithm}{S\arabic{algorithm}}
\setcounter{table}{0}
\setcounter{figure}{0}
\setcounter{algorithm}{0}
\twocolumn[{%
  \centering
  \Large
  \textbf{\thetitle}\\
  \vspace{0.5em}Supplementary Material \\
  \vspace{1.0em}
  \normalsize
  \includegraphics[width=\textwidth]{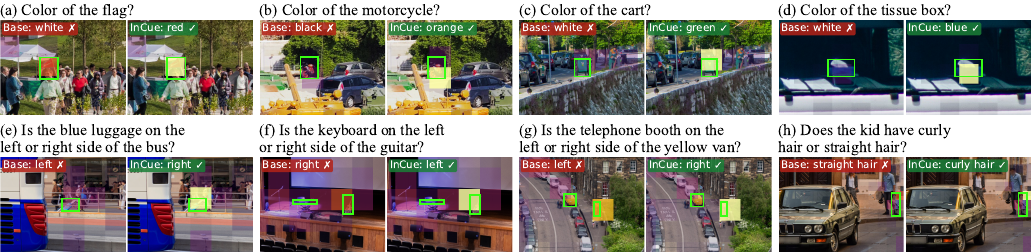}
  \captionof{figure}{\textbf{Examples corrected by InCue.}
  Eight V$^*$ questions that Qwen2.5-VL-7B answers wrongly, shown before (left) and after six InCue steps (right).
  Heat maps show attention over visual tokens, green boxes the ground-truth targets,
  and top-left labels the predictions (green correct, red wrong).
  InCue moves the attention onto the queried object and corrects the answer.}
  \label{fig:rescue_pairs}
  \vspace{1.2em}
}]
\appendix

\noindent Fig.~\ref{fig:rescue_pairs} shows V$^*$ questions that InCue corrects, in addition to Fig.~\ref{fig:rescue_steps}.
The rest of this supplementary material is organized as follows.
\begin{itemize}
\item Sec.~\ref{sec:supp_fixed}: results with one fixed setting per method and with settings chosen on held-out benchmarks.
\item Sec.~\ref{sec:supp_oracle}: the ground-truth-box oracle and controls on the target, the objective and the evidence.
\item Sec.~\ref{sec:supp_crop}: per-benchmark results of the crop-based comparison of Tab.~\ref{tab:crop}, including the native budget.
\item Sec.~\ref{sec:supp_cost}: computational cost, and a confidence gate that keeps 92\% of InCue's gain at one third of the optimization cost.
\item Sec.~\ref{sec:supp_diag}: the diagnostics of Fig.~\ref{fig:diag} across budgets, benchmarks and models, and raising attention logits instead of copying tokens.
\item Sec.~\ref{sec:supp_comp}: what Spatial Cue Coupling, Regional Cue Refocusing and the cue refresh do to the spatial target,
and a one-shot attention bias toward the same target.
\item Sec.~\ref{sec:supp_gen}: free-text answers instead of option letters.
\item Sec.~\ref{sec:supp_impl}: implementation details (hyper-parameters, algorithm, benchmarks and baselines).
\end{itemize}

\section{Results with One Fixed Setting per Method}
\label{sec:supp_fixed}
Tab.~\ref{tab:main} shows every method at its best setting and step.
Tab.~\ref{tab:supp_fixed} instead fixes one setting per method for all benchmarks and budgets.
InCue uses the method of Sec.~3 ($J$ with RCR, $\rho{=}4$, $\gamma{=}2$, cues and target refreshed at every step)
with one step size per base model ($\eta{=}1.92$ for Qwen2-VL, Qwen2.5-VL and InternVL3.5, $\eta{=}3.84$ for InternVL3)
and reads out the answer after $K{=}6$ steps.
AIF uses its default ratio $r{=}0.6$ with the shared readout, and TTRV adapts on a disjoint fold and is read out after its second epoch
(Sec.~\ref{sec:supp_baselines}).

InCue keeps the best average accuracy in all nine settings, with a mean gain of 2.3\% over the base model.
Across the evaluated model--budget--benchmark combinations, InCue improves over the base model in 24 of 27 comparisons, outperforms AIF in 21 of 24, and outperforms TTRV in all 8.

\begin{table}[t]
\centering
\caption{\textbf{Comparison with one fixed setting per method.} Same layout as Tab.~\ref{tab:main}.
InCue: one step size per base model, answer read out after $K{=}6$ steps; AIF: $r{=}0.6$;
TTRV: adaptation on a disjoint fold, second epoch. $^\dagger$Closest run without cue refresh, at step 6. Subscripts give
the gain over the base model.}
\label{tab:supp_fixed}
\renewcommand{\arraystretch}{1.1}
\setlength{\tabcolsep}{2.6pt}
\begin{adjustbox}{max width=\columnwidth}
\begin{tabular}{@{}c l c c c c@{}}
\toprule
Tokens & Method & V$^*$ & HR-4K & MME-RW & Avg. \\
\midrule
 \multirow{3}{*}{\shortstack{$\le$16K (native)}} & Qwen2.5-VL-7B & 79.1 & 71.8 & 52.7 & 67.8 \\
  & +\,AIF ($r{=}0.6$) & 78.0 & 72.5 & 53.4 & 68.0 \\
  & \cellcolor{oursbg}+\,\textbf{InCue} ($K{=}6$) & \cellcolor{oursbg}\textbf{81.7}\gain{+2.6} & \cellcolor{oursbg}\textbf{72.9}\gain{+1.1} & \cellcolor{oursbg}\textbf{55.0}\gain{+2.3} & \cellcolor{oursbg}\textbf{69.8}\gain{+2.0} \\
\midrule
 \multirow{4}{*}{\shortstack{2048}} & Qwen2.5-VL-7B & 68.6 & 64.0 & 48.6 & 60.4 \\
  & +\,AIF ($r{=}0.6$) & 71.7 & 66.1 & 48.3 & 62.0 \\
  & +\,TTRV & 68.6 & -- & -- & -- \\
  & \cellcolor{oursbg}+\,\textbf{InCue} ($K{=}6$) & \cellcolor{oursbg}\textbf{72.3}\gain{+3.7} & \cellcolor{oursbg}64.9\gain{+0.9} & \cellcolor{oursbg}\textbf{50.6}\gain{+2.0} & \cellcolor{oursbg}\textbf{62.6}\gain{+2.2} \\
\midrule
 \multirow{4}{*}{\shortstack{1024}} & Qwen2.5-VL-7B & 70.7 & 61.9 & 46.6 & 59.7 \\
  & +\,AIF ($r{=}0.6$) & 72.8 & 63.0 & 49.2 & 61.7 \\
  & +\,TTRV & 71.6 & 61.9 & 46.6 & 60.0 \\
  & \cellcolor{oursbg}+\,\textbf{InCue} ($K{=}6$) & \cellcolor{oursbg}\textbf{75.4}\gain{+4.7} & \cellcolor{oursbg}\textbf{63.4$^\dagger$}\gain{+1.5} & \cellcolor{oursbg}\textbf{49.9$^\dagger$}\gain{+3.3} & \cellcolor{oursbg}\textbf{62.9}\gain{+3.2} \\
\midrule
 \multirow{4}{*}{\shortstack{576}} & Qwen2.5-VL-7B & 61.3 & 57.3 & 42.6 & 53.7 \\
  & +\,AIF ($r{=}0.6$) & 62.8 & 58.4 & 43.2 & 54.8 \\
  & +\,TTRV & 62.5 & 58.2 & 42.7 & 54.4 \\
  & \cellcolor{oursbg}+\,\textbf{InCue} ($K{=}6$) & \cellcolor{oursbg}\textbf{64.4}\gain{+3.1} & \cellcolor{oursbg}\textbf{59.6$^\dagger$}\gain{+2.4} & \cellcolor{oursbg}\textbf{45.7}\gain{+3.1} & \cellcolor{oursbg}\textbf{56.6}\gain{+2.9} \\
\midrule
 \multirow{2}{*}{\shortstack{576}} & Qwen2-VL-7B & 56.0 & 59.9 & 46.0 & 54.0 \\
  & \cellcolor{oursbg}+\,\textbf{InCue} ($K{=}6$) & \cellcolor{oursbg}\textbf{60.2$^\dagger$}\gain{+4.2} & \cellcolor{oursbg}59.3\gain{$-$0.6} & \cellcolor{oursbg}\textbf{47.9}\gain{+1.9} & \cellcolor{oursbg}\textbf{55.8}\gain{+1.8} \\
\midrule
 \multirow{3}{*}{\shortstack{$\le$3328 (12 tiles)}} & InternVL3-8B & 70.7 & 71.3 & 55.1 & 65.7 \\
  & +\,AIF ($r{=}0.6$) & 74.3 & 72.3 & 57.1 & 67.9 \\
  & \cellcolor{oursbg}+\,\textbf{InCue} ($K{=}6$) & \cellcolor{oursbg}\textbf{77.5}\gain{+6.8} & \cellcolor{oursbg}70.9\gain{$-$0.4} & \cellcolor{oursbg}\textbf{57.4}\gain{+2.3} & \cellcolor{oursbg}\textbf{68.6}\gain{+2.9} \\
\midrule
 \multirow{3}{*}{\shortstack{$\le$2304 (8 tiles)}} & InternVL3-8B & 69.1 & 62.3 & 55.8 & 62.4 \\
  & +\,AIF ($r{=}0.6$) & 72.8 & 63.1 & 56.5 & 64.1 \\
  & \cellcolor{oursbg}+\,\textbf{InCue} ($K{=}6$) & \cellcolor{oursbg}\textbf{75.9}\gain{+6.8} & \cellcolor{oursbg}\textbf{63.9}\gain{+1.6} & \cellcolor{oursbg}\textbf{57.6}\gain{+1.8} & \cellcolor{oursbg}\textbf{65.8}\gain{+3.4} \\
\midrule
 \multirow{4}{*}{\shortstack{256 (1 tile)}} & InternVL3-8B & 59.2 & 57.3 & 46.6 & 54.3 \\
  & +\,AIF ($r{=}0.6$) & 60.2 & 57.6 & 46.5 & 54.8 \\
  & +\,TTRV & 58.8 & -- & -- & -- \\
  & \cellcolor{oursbg}+\,\textbf{InCue} ($K{=}6$) & \cellcolor{oursbg}\textbf{61.3$^\dagger$}\gain{+2.1} & \cellcolor{oursbg}\textbf{58.5}\gain{+1.3} & \cellcolor{oursbg}46.4\gain{$-$0.2} & \cellcolor{oursbg}\textbf{55.4}\gain{+1.1} \\
\midrule
 \multirow{3}{*}{\shortstack{256 (1 tile)}} & InternVL3.5-8B & 51.3 & 51.6 & 42.8 & 48.6 \\
  & +\,AIF ($r{=}0.6$) & 50.8 & 49.3 & 42.8 & 47.6 \\
  & \cellcolor{oursbg}+\,\textbf{InCue} ($K{=}6$) & \cellcolor{oursbg}\textbf{53.4}\gain{+2.1} & \cellcolor{oursbg}\textbf{52.3}\gain{+0.6} & \cellcolor{oursbg}\textbf{43.1}\gain{+0.3} & \cellcolor{oursbg}\textbf{49.6}\gain{+1.0} \\
\bottomrule
\end{tabular}
\end{adjustbox}
\end{table}

\noindent\textbf{Sensitivity to the setting.}
The gains do not depend on one configuration.
Choosing InCue's setting and step on a random half of each benchmark and evaluating on the other half
(2{,}000 random splits) gives a mean gain of 2.1\%, close to the fixed setting.
When the settings of AIF and TTRV are chosen in the same way, InCue leads AIF by 1.6\% (in 18 of 20 cells) and TTRV by 2.3\% (in 7 of 8).

\noindent\textbf{Held-out benchmarks.}
A setting chosen on two benchmarks also works on the third.
We choose InCue's setting and step on two of the three benchmarks and apply them unchanged to the third,
which plays no part in the choice.
On the held-out benchmark, InCue improves the base model in 14 of the 15 cases
(five pairs of base model and budget; Tab.~\ref{tab:supp_lobo}), by 2.6\% on average.

\begin{table}[t]
\centering
\caption{\textbf{Held-out benchmarks.} InCue's setting and step are chosen on the two other benchmarks
and applied unchanged to the benchmark of each column. Entries give the gain (\%) over the base model.}
\label{tab:supp_lobo}
\setlength{\tabcolsep}{5pt}
\begin{adjustbox}{max width=\columnwidth}
\begin{tabular}{@{}l l ccc@{}}
\toprule
Model & Tokens & V$^*$ & HR-4K & MME-RW \\
\midrule
\multirow{2}{*}{Qwen2.5-VL-7B} & $\le$16K (native) & +2.6 & +1.1 & +1.7 \\
 & 2048 & +4.7 & +1.9 & +3.1 \\
\midrule
\multirow{2}{*}{InternVL3-8B} & $\le$3328 (12 tiles) & +7.3 & $-$0.5 & +2.5 \\
 & $\le$2304 (8 tiles) & +8.4 & +1.8 & +2.4 \\
\midrule
InternVL3.5-8B & 256 (1 tile) & +1.6 & +0.8 & +0.3 \\
\bottomrule
\end{tabular}
\end{adjustbox}
\end{table}

\noindent\textbf{Significance.}
Tab.~\ref{tab:supp_sig} counts the answers that InCue corrects and breaks in each setting.
For Qwen2.5-VL-7B the gain is significant at every budget: InCue corrects 176 answers of the base model.
For InternVL3-8B with multiple tiles it is also significant (297 corrected).
Over all 22 table cells with stored answers, InCue corrects 741 answers.

\noindent\textbf{Paired comparison with AIF and TTRV.}
Since all methods answer the same questions, we also pair InCue with each baseline per question, each method as reported in Tab.~\ref{tab:supp_fixed},
over the cells where both store per-question outputs (20 for AIF, 4 for TTRV).
InCue is right and AIF wrong on 785 answers,
a significant lead of 1.0\% (2.5\% on V$^*$ and 1.1\% on MME-RW).
On these cells, InCue gains 1.6\% over the base model and AIF 0.6\%.
InCue also leads TTRV by 3.1\%, with a similar lead for each of TTRV's three seeds.

\begin{table}[t]
\centering
\caption{\textbf{Paired comparison with the base model under the fixed setting} of Tab.~\ref{tab:supp_fixed},
over all questions with stored per-question answers.
Corrected: wrong $\rightarrow$ right; broken: right $\rightarrow$ wrong.
$p$ is the chance of a gap at least this large between the two counts if InCue corrected and broke answers equally often.}
\label{tab:supp_sig}
\setlength{\tabcolsep}{4pt}
\begin{adjustbox}{max width=\columnwidth}
\begin{tabular}{@{}l c c c c@{}}
\toprule
Model & Tokens & Corrected & Broken & $p$ \\
\midrule
\multirow{4}{*}{Qwen2.5-VL-7B} & $\le$16K & 54 & 16 & $6\times10^{-6}$ \\
 & 2048 & 65 & 30 & $4\times10^{-4}$ \\
 & 1024 & 11 & 2 & $0.02$ \\
 & 576 & 46 & 12 & $8\times10^{-6}$ \\
\midrule
\multirow{2}{*}{InternVL3-8B} & $\le$3328 & 150 & 116 & $0.04$ \\
 & $\le$2304 & 147 & 102 & $0.005$ \\
\bottomrule
\end{tabular}
\end{adjustbox}
\end{table}

\section{Ground-Truth-Box Oracle and Controls}
\label{sec:supp_oracle}
\noindent\textbf{Oracle.}
To measure what a perfect target would give, we replace InCue's spatial target by the ground-truth box of V$^*$
(a target spread over the tokens inside the box, fixed over the steps) and keep the optimization of Sec.~\ref{sec:tto}.
Tab.~\ref{tab:supp_oracle} compares the two after the same six steps, with InCue in the fixed setting of Tab.~\ref{tab:supp_fixed}.
Without any annotation, InCue recovers 64--75\% of the oracle's gain on both model families,
for example +4.7\% of +6.3\% at 1024 tokens and +6.8\% of +9.9\% on InternVL3-8B.

\noindent\textbf{Where the target points matters.}
We move the target away from the evidence by randomly permuting the positions of the oracle target (576 tokens).
The accuracy then returns to the base level (60.7--61.8\% against 61.3\%), whereas the true box gives 66.0--66.5\%.

\noindent\textbf{Minimizing the entropy directly.}
InCue uses the gradient of the predictive entropy only to locate where the prediction is sensitive,
whereas entropy-based test-time adaptation lowers the entropy itself.
We test the latter by updating the same visual tokens with gradient descent on the option entropy for six steps,
with the step size matched to InCue's first update (and 0.25 and 4 times that size).
The model becomes more confident (on V$^*$ at 576 tokens, the mean entropy falls from 0.53 to 0.24),
but with any of the three step sizes at most 2 of the 191 V$^*$ answers change.
The accuracy stays within 1.1\% of the base level
(+0.0\%, +0.5\% and +1.0\% on V$^*$ at 576, 1024 and 2048 tokens and +1.1\% on MME-RW at 576 tokens; top of Tab.~\ref{tab:supp_ctrl}),
whereas InCue gains 3.1\%, 4.7\%, 3.7\% and 3.1\% in the same settings.
On MME-RW, the entropy falls from 0.74 to 0.39, while the correct-answer margin
(the logit of the correct option minus the strongest wrong one) drops by 0.22;
InCue barely changes the entropy (0.74 to 0.75) and raises this margin by 0.26.
Lowering the uncertainty alone does not give the gain.

\begin{table}[t]
\centering
\caption{\textbf{InCue against the ground-truth-box oracle} on V$^*$. The oracle steers the attention toward the ground-truth box.
Both are read out after $K{=}6$ steps, and InCue uses the fixed setting of Tab.~\ref{tab:supp_fixed}.
Subscripts give the gain over the base model, and Recovered is InCue's gain divided by the oracle's gain.}
\label{tab:supp_oracle}
\setlength{\tabcolsep}{3.5pt}
\begin{adjustbox}{max width=\columnwidth}
\begin{tabular}{@{}l c c c c c@{}}
\toprule
Model & Tokens & Base & \cellcolor{oursbg}InCue & Oracle & Recovered \\
\midrule
\multirow{4}{*}{Qwen2.5-VL-7B} & $\le$16K & 79.1 & \cellcolor{oursbg}81.7\gain{+2.6} & 82.7\gain{+3.7} & 71\% \\
 & 2048 & 68.6 & \cellcolor{oursbg}72.3\gain{+3.7} & 74.3\gain{+5.8} & 64\% \\
 & 1024 & 70.7 & \cellcolor{oursbg}75.4\gain{+4.7} & 77.0\gain{+6.3} & 75\% \\
 & 576  & 61.3 & \cellcolor{oursbg}64.4\gain{+3.1} & 65.4\gain{+4.2} & 75\% \\
\midrule
\multirow{2}{*}{InternVL3-8B} & $\le$3328 & 70.7 & \cellcolor{oursbg}77.5\gain{+6.8} & 80.6\gain{+9.9} & 68\% \\
 & $\le$2304 & 69.1 & \cellcolor{oursbg}75.9\gain{+6.8} & 79.1\gain{+9.9} & 68\% \\
\bottomrule
\end{tabular}
\end{adjustbox}
\end{table}

\noindent\textbf{InCue's corrections rest on the evidence.}
After InCue's six steps, we mask the visual tokens that overlap a ground-truth box
(4.0, 5.6 and 8.4 tokens on average at 576, 1024 and 2048 tokens) by replacing them with the mean of the other tokens,
and, as a control, the same number of tokens shifted to a position outside the boxes.
Masking the evidence lowers InCue's accuracy from 64.4\% to 37.7\% at 576 tokens, from 75.4\% to 44.5\% at 1024
and from 72.3\% to 44.5\% at 2048,
whereas masking the control tokens neither loses nor gains a correct answer (bottom of Tab.~\ref{tab:supp_ctrl}).

\begin{table}[t]
\centering
\caption{\textbf{Entropy minimization and evidence masking} (Qwen2.5-VL-7B, accuracy in \%).
Top: gradient descent on the option entropy over the visual tokens for six steps, with the step size matched to InCue's first update.
Bottom (V$^*$, which has ground-truth boxes): after InCue's six steps, the tokens that overlap a ground-truth box (evidence)
or the same number of tokens outside the boxes (control) are replaced by the mean of the other tokens.
Subscripts give the gain over the base model.}
\label{tab:supp_ctrl}
\setlength{\tabcolsep}{3pt}
\begin{adjustbox}{max width=\columnwidth}
\begin{tabular}{@{}l c c c c@{}}
\toprule
 & \multicolumn{3}{c}{V$^*$} & MME-RW \\
\cmidrule(lr){2-4}\cmidrule(l){5-5}
Tokens & 576 & 1024 & 2048 & 576 \\
\midrule
Base & 61.3 & 70.7 & 68.6 & 42.6 \\
Entropy minimization & 61.3\gain{+0.0} & 71.2\gain{+0.5} & 69.6\gain{+1.0} & 43.7\gain{+1.1} \\
\cellcolor{oursbg}InCue & \cellcolor{oursbg}\textbf{64.4}\gain{+3.1} & \cellcolor{oursbg}\textbf{75.4}\gain{+4.7} & \cellcolor{oursbg}\textbf{72.3}\gain{+3.7} & \cellcolor{oursbg}\textbf{45.7}\gain{+3.1} \\
\midrule
Evidence masked & 37.7 & 44.5 & 44.5 & -- \\
Control masked & 64.4 & 75.4 & 72.3 & -- \\
\bottomrule
\end{tabular}
\end{adjustbox}
\end{table}

\section{Crop-Based Comparison per Benchmark}
\label{sec:supp_crop}

\begin{table*}[t]
\centering
\caption{\textbf{Region cues under the same crop rule, per benchmark} (Qwen2.5-VL-7B, 576 tokens plus the crop, $K{=}0$: no optimization).
The bottom of Tab.~\ref{tab:crop} averages them over the three benchmarks. Windows are given by their side relative to the image;
boxes follow ViCrop's rule with the given base size in pixels.}
\label{tab:supp_cropgrid}
\setlength{\tabcolsep}{4pt}
\begin{adjustbox}{max width=\textwidth}
\begin{tabular}{@{}l ccccc ccccc ccccc@{}}
\toprule
 & \multicolumn{5}{c}{V$^*$} & \multicolumn{5}{c}{HR-4K} & \multicolumn{5}{c}{MME-RW} \\
\cmidrule(lr){2-6}\cmidrule(lr){7-11}\cmidrule(lr){12-16}
 & \multicolumn{2}{c}{Window} & \multicolumn{3}{c}{Box (px)} & \multicolumn{2}{c}{Window} & \multicolumn{3}{c}{Box (px)} & \multicolumn{2}{c}{Window} & \multicolumn{3}{c}{Box (px)} \\
Region cue & 1/8 & 1/2 & 336 & 448 & 672 & 1/8 & 1/2 & 336 & 448 & 672 & 1/8 & 1/2 & 336 & 448 & 672 \\
\midrule
ViCrop & 62.3 & 63.9 & 64.4 & 62.3 & 64.4 & 58.8 & 64.2 & 58.0 & 60.5 & 64.2 & 42.7 & 47.1 & 44.7 & 45.0 & 46.2 \\
ViCrop (tuned) & 70.7 & 74.3 & 70.7 & 75.9 & 73.3 & 69.5 & 67.2 & 67.5 & 70.0 & 71.2 & 48.3 & 52.5 & 49.0 & 49.4 & 49.2 \\
HiDe & 68.6 & 69.1 & 63.9 & 69.6 & 72.3 & 58.8 & 67.5 & 63.4 & 64.6 & 66.1 & 43.7 & 50.1 & 46.8 & 46.5 & 48.3 \\
EGG & 69.1 & 67.5 & 71.7 & 72.3 & 72.3 & 64.1 & 69.0 & 62.9 & 66.4 & 68.5 & 48.4 & 52.6 & 49.0 & 49.8 & 51.1 \\
\cellcolor{oursbg}InCue ($J$) & \cellcolor{oursbg}\textbf{71.7} & \cellcolor{oursbg}73.8 & \cellcolor{oursbg}\textbf{71.7} & \cellcolor{oursbg}74.3 & \cellcolor{oursbg}\textbf{74.9} & \cellcolor{oursbg}67.9 & \cellcolor{oursbg}\textbf{69.1} & \cellcolor{oursbg}66.2 & \cellcolor{oursbg}\textbf{70.5} & \cellcolor{oursbg}\textbf{71.6} & \cellcolor{oursbg}\textbf{50.3} & \cellcolor{oursbg}\textbf{53.3} & \cellcolor{oursbg}\textbf{49.7} & \cellcolor{oursbg}\textbf{50.8} & \cellcolor{oursbg}\textbf{51.6} \\
\bottomrule
\end{tabular}
\end{adjustbox}
\end{table*}

\begin{table}[t]
\centering
\caption{\textbf{Share of V$^*$ questions (\%) whose crop contains the target}, \ie, the centers of all ground-truth boxes,
for the region cues and crop rules of Tab.~\ref{tab:supp_cropgrid} (Qwen2.5-VL-7B, 576 tokens).}
\label{tab:supp_crophit}
\setlength{\tabcolsep}{5pt}
\begin{tabular}{@{}l cc ccc@{}}
\toprule
 & \multicolumn{2}{c}{Window} & \multicolumn{3}{c}{Box (px)} \\
\cmidrule(lr){2-3}\cmidrule(l){4-6}
Region cue & 1/8 & 1/2 & 336 & 448 & 672 \\
\midrule
ViCrop & 7.3 & 48.7 & 13.6 & 19.4 & 31.4 \\
ViCrop (tuned) & 49.2 & 78.0 & 59.7 & 62.8 & 70.2 \\
HiDe & 30.9 & 68.1 & 46.1 & 55.0 & 60.2 \\
EGG & 44.5 & 74.9 & 49.7 & 56.0 & 66.0 \\
\cellcolor{oursbg}InCue ($J$) & \cellcolor{oursbg}\textbf{53.4} & \cellcolor{oursbg}\textbf{83.2} & \cellcolor{oursbg}\textbf{62.3} & \cellcolor{oursbg}\textbf{65.4} & \cellcolor{oursbg}\textbf{72.8} \\
\bottomrule
\end{tabular}
\end{table}

\begin{table}[t]
\centering
\caption{\textbf{Crop-based methods with InCue at the native budget} (V$^*$, Qwen2.5-VL-7B, $\le$16K tokens plus the crop).
InCue uses its native-resolution setting of Tab.~\ref{tab:main} (fixed target with RCR, $\eta{=}3.84$)
and is read out at step 7 for all three crops. Subscripts give the gain over the crop alone.}
\label{tab:supp_cropnative}
\setlength{\tabcolsep}{6pt}
\begin{tabular}{@{}l c c@{}}
\toprule
Crop & Crop alone & \cellcolor{oursbg}+\,\textbf{InCue} \\
\midrule
ViCrop~\cite{zhang2025vicrop} & 78.0 & \cellcolor{oursbg}\textbf{80.6}\gain{+2.6} \\
HiDe~\cite{liu2026hide} & 91.6 & \cellcolor{oursbg}\textbf{92.7}\gain{+1.0} \\
Crop selected by $J$ & 84.3 & \cellcolor{oursbg}\textbf{86.4}\gain{+2.1} \\
\bottomrule
\end{tabular}
\end{table}

\noindent\textbf{Region cues under the same crop rule.}
The bottom of Tab.~\ref{tab:crop} averages the three benchmarks; Tab.~\ref{tab:supp_cropgrid} gives each of them.
With the crop rule fixed and no optimization, our joint response $J$ is the best region cue
in 11 of the 15 benchmark--rule pairs (one of them tied), including all five on \mbox{MME-RW}, and the second best in the other four.

\noindent\textbf{Why $J$ selects better crops.}
A crop can only help if it contains the evidence.
Tab.~\ref{tab:supp_crophit} measures this on V$^*$ with the ground-truth boxes, for the crops of Tab.~\ref{tab:supp_cropgrid}.
Under each of the five crop rules, the crop selected by $J$ contains the target most often,
\eg, for 83.2\% of the questions with the half-image window,
against 78.0\% for the tuned ViCrop, 74.9\% for EGG and 68.1\% for HiDe.

\noindent\textbf{Native budget.}
InCue still helps on top of a crop when the input already has native resolution (Tab.~\ref{tab:supp_cropnative}).
On V$^*$, it raises ViCrop from 78.0\% to 80.6\%, HiDe from 91.6\% to 92.7\% and the crop selected by $J$ from 84.3\% to 86.4\%,
with one setting and one read-out step for all three crops.
On 200 HR-4K questions (one option order each), InCue with the fixed setting of Tab.~\ref{tab:supp_fixed}
raises the crop selected by $J$ from 74.5\% to 76.0\% and breaks no correct answer.

\section{Computational Cost}
\label{sec:supp_cost}
InCue adds no visual tokens, so its memory stays close to that of the base model.
The peak is within 2\,GiB of the base model up to 2048 tokens, and native resolution fits on a single 80\,GB GPU
(Tab.~\ref{tab:supp_cost}), where TTRV runs out of memory on four.
The extra cost of InCue is time, since each step runs one forward and two backward passes through the LLM.
TTRV instead adapts before testing, which takes \mbox{21--26}~minutes on four GPUs per run.

\noindent\textbf{Confidence gate.}
InCue does not have to run on every question.
The confidence of the base model is known before any optimization,
and InCue's gains come mostly from low-confidence questions (Tab.~\ref{tab:hard}).
We therefore run InCue only on the third of the questions with the smallest confidence gap of the base model,
selected by the label-free rule of Tab.~\ref{tab:hard}, and keep the base answer elsewhere.
Under the fixed setting of Tab.~\ref{tab:supp_fixed}, this gate keeps 92\% of InCue's gain
(Tab.~\ref{tab:supp_gate}).
On V$^*$ it keeps 96\% of the gain (the gated third alone improves by 11.7\%),
and on HR-4K it is even better than running InCue on every question.

Tab.~\ref{tab:supp_cost} gives the absolute numbers at native resolution, where the input cannot be enlarged further.
AIF lowers the accuracy there and TTRV does not fit, whereas InCue with the gate gains 2.6\%.
At this resolution the gate keeps the accuracy of running InCue on every question at every step (Tab.~\ref{tab:supp_steps}),
so $K$ trades accuracy against latency.

\begin{table}[t]
\centering
\caption{\textbf{Latency, peak memory and accuracy} on V$^*$ at native resolution (Qwen2.5-VL-7B, one NVIDIA A100-80GB,
batch size 1, our FP32 research implementation; latency per question includes the vision encoder).
InCue uses the fixed setting of Tab.~\ref{tab:supp_fixed} with the confidence gate of Tab.~\ref{tab:supp_gate},
so its latency is the base latency plus one third of the optimization time.}
\label{tab:supp_cost}
\renewcommand{\arraystretch}{1.08}
\setlength{\tabcolsep}{4pt}
\begin{adjustbox}{max width=\columnwidth}
\begin{tabular}{@{}l c c c@{}}
\toprule
Method & Latency (s) & Memory (GiB) & Acc. (\%) \\
\midrule
Qwen2.5-VL-7B & 6.2 & 35.9 & 79.1 \\
+\,AIF ($r{=}0.6$) & 10.5 & 32.8 & 78.0 \\
+\,TTRV & \multicolumn{3}{c}{out of memory on four 80\,GB GPUs} \\
\cellcolor{oursbg}+\,\textbf{InCue} ($K{=}6$, gate) & \cellcolor{oursbg}52.5 & \cellcolor{oursbg}45.7 & \cellcolor{oursbg}\textbf{81.7} \\
\bottomrule
\end{tabular}
\end{adjustbox}
\end{table}

\begin{table}[t]
\centering
\caption{\textbf{Confidence gate.} InCue under the fixed setting of Tab.~\ref{tab:supp_fixed}, run on all questions
or only on the third with the smallest confidence gap of the base model (label-free rule of Tab.~\ref{tab:hard}),
pooled over the base models and budgets with stored confidences (Cells). Gains are over the base model on all questions,
and Kept is the gain with the gate divided by the gain without it.}
\label{tab:supp_gate}
\setlength{\tabcolsep}{4pt}
\begin{adjustbox}{max width=\columnwidth}
\begin{tabular}{@{}l c c c c c@{}}
\toprule
Benchmark & Cells & Answers & \shortstack{Gain (\%)\\all questions} & \shortstack{Gain (\%)\\gated third} & Kept \\
\midrule
V$^*$   & 6 & 1{,}146 & +4.10 & +3.93 & 96\% \\
HR-4K   & 7 & 5{,}600 & +0.68 & +0.77 & 113\% \\
MME-RW  & 8 & 8{,}368 & +1.64 & +1.40 & 85\% \\
\midrule
All     & 21 & 15{,}114 & +1.47 & +1.36 & 92\% \\
\bottomrule
\end{tabular}
\end{adjustbox}
\end{table}

\begin{table}[t]
\centering
\caption{\textbf{Accuracy against latency over InCue steps with the confidence gate} on V$^*$ at native resolution
(Qwen2.5-VL-7B, same GPU and latency rule as Tab.~\ref{tab:supp_cost}). Step 0 is the base model.}
\label{tab:supp_steps}
\setlength{\tabcolsep}{3.2pt}
\begin{adjustbox}{max width=\columnwidth}
\begin{tabular}{@{}l ccccccc@{}}
\toprule
Step $k$ & 0 & 1 & 2 & 3 & 4 & 5 & 6 \\
\midrule
Acc. (\%) & 79.1 & 80.6 & 80.6 & 80.1 & 81.2 & 81.7 & 81.7 \\
Latency (s) & 6.2 & 17.4 & 24.4 & 31.4 & 38.5 & 45.5 & 52.5 \\
\bottomrule
\end{tabular}
\end{adjustbox}
\end{table}

\section{Evidence Is Already in the Model: Extended Diagnostics}
\label{sec:supp_diag}
Fig.~\ref{fig:diag} shows each diagnostic in one setting.
Here we repeat them across budgets, region sizes, datasets and models.

\noindent\textbf{Setup.}
Let $R$ be the question's region: the ground-truth box enlarged to a square window whose side is at least a quarter of the image,
unless stated otherwise. Against the base model we compare four ways of re-presenting $R$:
(i) appending the crop of $R$ from the original image, re-encoded;
(ii) appending the crop of $R$ from the downsampled input that the model has already seen, re-encoded at the same resolution;
(iii) appending a copy of the cached tokens of $R$, resampled to the grid of the crop, without re-encoding;
(iv) appending nothing and raising the attention logits of the tokens of $R$ by $\log(1{+}m_i)$
at all text positions and layers, where $m_i$ is the number of copies of token $i$ in (iii).
As controls, a same-size region that does not overlap $R$ replaces $R$,
and copies of whole-image tokens replace those of $R$ in (iii).

\noindent\textbf{No new detail is needed.}
Fig.~\ref{fig:diag}a holds in all 32 settings in which the original image has more pixels than the input
(Qwen2.5-VL-7B and LLaVA-1.5-7B; V$^*$, HR-4K and MME-RW; 576 to 2048 tokens;
ground-truth windows of three sizes and attention-selected windows of two sizes).
In each of them, the crop from the already-seen input improves the base model, by a median of 9.0\%,
and recovers a median of 85\% of the gain of the crop from the original image.
A same-size crop of another region adds at most 1.6\% on V$^*$, so the gain comes from the region itself.

\noindent\textbf{Attention alone carries part of the gain.}
Tab.~\ref{tab:supp_diag} repeats Fig.~\ref{fig:diag}b on V$^*$ at all four budgets.
Copying the region's cached tokens improves the base model at every budget (+1.0 to +4.7\%),
whereas copying as many tokens from the whole image or from another region gives at most +1.6\%.
Raising the attention logits of the same tokens, with nothing appended, gives +2.6 to +6.3\%,
more than the copy at every budget.
Over all 27 settings with a ground-truth window (both models; V$^*$, Visual Genome and COCO),
raising the logits improves the base model in 25, with a median gain of 3.5\%
(copy: 2.6\%; whole-image copy: 0.5\%).
Part of the gain of re-presenting the region therefore needs no additional visual input, and InCue is built on this part.

\begin{table}[t]
\centering
\caption{\textbf{Copying cached tokens and raising attention logits} for the ground-truth region
(V$^*$, Qwen2.5-VL-7B, BF16 inference). Base gives the accuracy (\%), the other columns the gain over it.
Copies append cached tokens of the region, of the whole image or of another region without re-encoding;
the last column appends nothing and only raises the attention logits of the region's tokens.}
\label{tab:supp_diag}
\setlength{\tabcolsep}{4pt}
\begin{adjustbox}{max width=\columnwidth}
\begin{tabular}{@{}c c ccc c@{}}
\toprule
 & & \multicolumn{3}{c}{Copied tokens} & \multirow{2}{*}{\shortstack{Raised\\logits}} \\
\cmidrule(lr){3-5}
Tokens & Base & region & image & other & \\
\midrule
576      & 62.8 & +1.0 & +0.5 & $-1.0$ & +2.6 \\
1024     & 70.7 & +4.7 & +1.6 & +1.0   & +6.3 \\
2048     & 68.6 & +4.2 & +1.0 & +0.5   & +5.8 \\
$\le$16K & 79.1 & +3.7 & $-1.0$ & $-2.1$ & +4.2 \\
\bottomrule
\end{tabular}
\end{adjustbox}
\end{table}

\section{Analysis of the Components}
\label{sec:supp_comp}
Tab.~\ref{tab:ablation} ablates InCue's components by accuracy.
Here we measure what each component does to the spatial target and to the attention.
The ground-truth boxes of V$^*$ are used only for measurement (Qwen2.5-VL-7B, all 191 questions).

\noindent\textbf{Spatial Cue Coupling.}
A spatial target is useful only if it points at the evidence.
For each map, we select the half-image window with the largest mass among the candidates of RCR
and check whether it contains the centers of all ground-truth boxes (left of Tab.~\ref{tab:supp_window}).
The window selected on the joint response $J$ contains the target more often than the window selected on the attention cue $A$
or on the sensitivity cue $S$ at every budget, \eg, for 83.2\% of the questions against 78.5\% and 72.8\% at 576 tokens.
The lead remains in all eight comparisons when each cue is sharpened to the entropy of $J$,
so it does not come from $J$ being a sharper map.
Coupling also puts more of the target on the evidence.
$J$ places 11.9--26.5\% of its mass inside the ground-truth boxes, against at most 4.8\% for either cue (right of Tab.~\ref{tab:supp_window}).

\begin{table}[t]
\centering
\caption{\textbf{What each component does to the spatial target} (V$^*$, Qwen2.5-VL-7B; ground-truth boxes only for measurement).
Left: share of questions (\%) for which the half-image window selected on a map contains the centers of all ground-truth boxes;
a random window does so for 28.2\%.
Right: share of each map's mass (\%) inside the ground-truth boxes, with the refreshed target taken at the last step;
a uniform map has 0.17\%.}
\label{tab:supp_window}
\setlength{\tabcolsep}{3pt}
\begin{adjustbox}{max width=\columnwidth}
\begin{tabular}{@{}l ccc ccccc@{}}
\toprule
 & \multicolumn{3}{c}{Window contains target} & \multicolumn{5}{c}{Mass inside the boxes} \\
\cmidrule(lr){2-4}\cmidrule(l){5-9}
Tokens & $A$ & $S$ & $J$ & $A$ & $S$ & $J$ & + RCR & + refresh \\
\midrule
576 & 78.5 & 72.8 & \textbf{83.2} & 2.4 & 2.4 & 11.9 & 13.9 & \textbf{15.6} \\
1024 & 82.2 & 81.2 & \textbf{83.2} & 3.0 & 2.8 & 14.4 & 17.0 & \textbf{18.6} \\
2048 & 87.4 & 82.2 & \textbf{89.0} & 4.2 & 3.6 & 23.5 & 26.4 & \textbf{30.2} \\
$\le$16K & 90.1 & 89.0 & \textbf{91.6} & 4.8 & 3.8 & 26.5 & 30.8 & \textbf{35.0} \\
\bottomrule
\end{tabular}
\end{adjustbox}
\end{table}

The better target also moves more attention onto the evidence during optimization.
After six steps with fixed targets at 2048 tokens, the attention inside the ground-truth boxes ends higher with $J$
than with $A$ alone for 128 of 191 questions, and higher than with $S$ alone for 135.
$J$ also raises the correct-answer margin by 0.26,
against 0.15 for $A$ alone.

\noindent\textbf{Regional Cue Refocusing.}
RCR amplifies $J$ inside the selected window $W^\star$, so it helps only if $W^\star$ contains the evidence.
This holds for 83.2--91.6\% of the questions (column $J$ on the left of Tab.~\ref{tab:supp_window}),
about three times the 28.2\% of a random window, and $W^\star$ covers 85--93\% of the box area on average.
With RCR, the target mass inside the boxes rises for 164, 166, 177 and 183 of the 191 questions at the four budgets,
\eg, from 26.5\% to 30.8\% at native resolution.
In accuracy, over all 44 pairs of Qwen2.5-VL-7B runs that differ only in RCR
(four budgets, three benchmarks, fixed and refreshed targets, six steps),
RCR is ahead in 21 pairs, tied in 20 and behind in 3.

\noindent\textbf{Cue refresh.}
The cues change while the visual tokens are updated.
Within six steps, the window selected on the current cues comes to differ from the one selected before the first step
for 43\%, 37\%, 38\% and 41\% of the questions at 576, 1024 and 2048 tokens and at native resolution,
which a target built once cannot follow.
Refreshing also moves the target further onto the evidence.
From the first to the last step, its mass inside the ground-truth boxes rises at every budget
(last column of Tab.~\ref{tab:supp_window}), for 122, 123, 134 and 146 of the 191 questions.
In accuracy, over all 27 pairs of runs of Qwen2.5-VL-7B that differ only in the refresh,
the refreshed target is ahead in 15 pairs, tied in 10 and behind in 2 after six steps,
and it is ahead in 6 and tied in 2 of the 8 pairs with twelve steps.

\begin{table}[t]
\centering
\caption{\textbf{One-shot attention bias toward InCue's target} (V$^*$, Qwen2.5-VL-7B, 576 tokens).
The bias uses the same target as InCue but optimizes nothing.
Median and best are over its twelve settings, and the best is chosen with the test labels.
InCue uses the fixed setting of Tab.~\ref{tab:supp_fixed}, read out after one step (first step) or six.
Subscripts give the gain over the base model.}
\label{tab:supp_bias}
\setlength{\tabcolsep}{6pt}
\begin{tabular}{@{}l c@{}}
\toprule
Method & Accuracy (\%) \\
\midrule
Base & 61.3 \\
Bias, median of 12 settings & 61.0\gain{$-$0.3} \\
Bias, fixed in advance ($\rho{=}4$, read-out) & 62.8\gain{+1.6} \\
Bias, best of 12 settings & 63.4\gain{+2.1} \\
InCue, first step & 63.4\gain{+2.1} \\
\cellcolor{oursbg}InCue & \cellcolor{oursbg}\textbf{64.4}\gain{+3.1} \\
\bottomrule
\end{tabular}
\end{table}

With all three components, the attention inside the ground-truth boxes increases for 76\%, 83\% and 88\% of the V$^*$ questions
at 576 and 2048 tokens and native resolution.

\noindent\textbf{Test-time optimization.}
A simpler use of InCue's spatial target is to bias the attention toward it without optimizing the visual tokens.
We add the target of the first step directly to the attention logits of every text position after the image,
in twelve settings (a window or a dense bias, three strengths each, in the read-out layers or in all layers).
On V$^*$ at 576 tokens, this one-shot bias gives only a weak gain that depends strongly on its form and strength (Tab.~\ref{tab:supp_bias}).
Only four of the twelve settings improve on the base model, the median of the twelve is 0.3\% below it,
and the strongest biases lower the accuracy by up to 22.5\%.
The setting fixed in advance, a window bias with the odds of RCR in the read-out layers, gains 1.6\%.
The best of the twelve settings, chosen with the test labels, gains 2.1\%,
which InCue already reaches after its first step with its one fixed setting.
After six steps, InCue is above all twelve settings.
The target alone thus already helps, and optimizing the visual tokens toward it doubles the gain of the bias fixed in advance.
Sec.~\ref{sec:supp_cost} gives the cost of the steps and a gate that runs them on a third of the questions.

\section{Generated Answers}
\label{sec:supp_gen}
InCue's gains are not tied to the option-letter readout.
Its optimization is unchanged, with the entropy and its gradient computed over the option letters;
only the final answer is read out as in AIF~\cite{liu2026aif}: the model generates a free-text answer from the adjusted visual tokens,
prompted to answer in a single word or phrase, and the answer is matched to the options.
The base model answers likewise from $U_0$.
After six steps without cue refresh, InCue raises the V$^*$ accuracy of Qwen2.5-VL-7B from 70.7\% to 74.3\% at 1024 tokens.

\section{Implementation Details}
\label{sec:supp_impl}

\subsection{Hyper-Parameters}
\label{sec:supp_hparams}
\noindent\textbf{Setting.}
Tab.~\ref{tab:supp_hparams} lists the hyper-parameters of InCue.
We aggregate attention from a fixed set of decoder layers, the read-out layers $\Lambda$. The same selection rule is used for every base model, without tuning:
the upper half of the decoder without its last two layers, i.e., layers $L/2{+}1$ to $L{-}2$ of a decoder with $L$ layers.
The attention cue averages the text-to-visual attention over all heads of these layers and over all text positions after the image.
RCR scans $13{\times}13$ windows whose sides are half of the image sides, placed at evenly spaced positions.
For multi-tile inputs of InternVL, every visual token, including those of the thumbnail, is mapped to its box in the original image,
so the windows and the target are defined on the image and not on the tiles.
The offset is updated by plain gradient descent without normalizing the gradient.
The LLM runs in FP32 and the vision encoder in BF16; the visual tokens are computed once and cached.

\begin{table}[t]
\centering
\caption{\textbf{Hyper-parameters of InCue.} $L$: decoder layers; $\Lambda$: read-out layers, used for attention aggregation;
$\eta$: step size; $\rho$: odds; $\gamma$: norm; $K$: steps.
This is the setting of Tab.~\ref{tab:supp_fixed}.}
\label{tab:supp_hparams}
\setlength{\tabcolsep}{4pt}
\begin{adjustbox}{max width=\columnwidth}
\begin{tabular}{@{}l c c c c c c@{}}
\toprule
Base model & $L$ & $\Lambda$ & $\eta$ & $\rho$ & $\gamma$ & $K$ \\
\midrule
Qwen2-VL-7B    & 28 & 15--26 & 1.92 & 4 & 2 & 6 \\
Qwen2.5-VL-7B  & 28 & 15--26 & 1.92 & 4 & 2 & 6 \\
InternVL3-8B   & 28 & 15--26 & 3.84 & 4 & 2 & 6 \\
InternVL3.5-8B & 36 & 19--34 & 1.92 & 4 & 2 & 6 \\
\bottomrule
\end{tabular}
\end{adjustbox}
\end{table}

\noindent\textbf{Search.}
Tab.~\ref{tab:main} reports the best setting and step that we ran in every cell, for InCue as for the baselines (Sec.~\ref{sec:supp_baselines}).
For InCue, the step size is searched from 0.24 to 15.36 in factors of two, together with the odds, the norm and the read-out step.
$\Lambda$ and the windows of RCR are never tuned, whereas ViCrop (tuned) chooses its attention layer and crop on each benchmark.
Sec.~\ref{sec:supp_fixed} removes this search for all methods and fixes the setting of Tab.~\ref{tab:supp_hparams}.

\subsection{Algorithm}
\label{sec:supp_alg}
Alg.~\ref{alg:supp_incue} summarizes InCue with the notation of Sec.~3.
Each step runs one forward pass, which gives the attention cue and the predictive entropy, and two backward passes,
one for the sensitivity cue and one for the update of the offset.
The model parameters are never updated, and the image is encoded only once.

\subsection{Benchmarks}
\label{sec:supp_benchmarks}
\noindent\textbf{Fine-grained benchmarks.}
\emph{V$^*$}~\cite{wu2024vstar}: all 191 questions (115 on attributes, 76 on spatial relations).
\emph{HR-Bench 4K}~\cite{wang2024hrbench}: all 800 questions (400 single-instance and 400 cross-instance perception).
\emph{MME-RealWorld-Lite}~\cite{zhang2024mmerealworld}: the perception split, 1{,}046 questions after removing duplicated rows
and five questions flagged by our data-overlap check
(autonomous driving 309, OCR 250, monitoring 241, remote sensing 150, diagrams and tables 96).
The eight general benchmarks of Tab.~\ref{tab:general} have 7{,}246 questions in total.

\noindent\textbf{Visual-token budgets.}
For Qwen2-VL and Qwen2.5-VL, an image is resized with its aspect ratio kept so that it gives at most $B$ visual tokens
($B \in \{576, 1024, 2048\}$); smaller images are never upsampled.
The native budget uses the default limit of the model, 16{,}384 tokens.
For InternVL, the image is split into at most 1, 8 or 12 tiles of $448{\times}448$ pixels with 256 tokens each,
plus a thumbnail when there are several tiles: at most 256, 2{,}304 and 3{,}328 tokens.

\noindent\textbf{Prompt and readout.}
Each question is given as one user turn with the image first, in the official chat template of the model:
the question, one line per option, and the instruction
``Answer with the option's letter from the given choices directly.''
The answer is the option letter with the highest logit at the first decoding step;
for InternVL3, the logits of a letter's token variants (e.g., with a leading space) are merged.

\begin{algorithm}[t]
\caption{InCue}
\label{alg:supp_incue}
\begin{algorithmic}[1]
\Require image $I$, prompt $x$, frozen MLLM with encoder $\mathcal{E}$; steps $K$, step size $\eta$, odds $\rho$, norm $\gamma$
\State $U_0 \gets \mathcal{E}(I)$, \ $\Delta U^{(0)} \gets 0$ \Comment{encode once, cache}
\For{$k = 0, \dots, K{-}1$}
  \State $U^{(k)} \gets U_0 + \Delta U^{(k)}$
  \State $A^{(k)},\, H \gets$ forward pass at $U^{(k)}$ \Comment{Eqs.~(2), (3)}
  \State $S^{(k)} \gets \mathrm{Norm}\big([\lVert \partial H / \partial u_i \rVert_2]_{i}\big)$ \Comment{Eq.~(4)}
  \State $J^{(k)} \gets \mathrm{Norm}\big(A^{(k)} \odot S^{(k)}\big)$ \Comment{Eq.~(5)}
  \State $W^{\star} \gets \arg\max_{W \in \mathcal{W}} \phi(W)$ \Comment{Eq.~(6)}
  \State $T^{(k)} \gets \mathrm{Norm}\big([J^{(k)}_i \rho^{\,\mathbb{I}[i \in W^{\star}]}]_{i}\big)$ \Comment{Eq.~(7)}
  \State $\mathcal{L}^{(k)} \gets -\log \lVert A^{(k)} \odot \stopgrad[T^{(k)}] \rVert_{\gamma}$ \Comment{Eq.~(8)}
  \State $\Delta U^{(k+1)} \gets \Delta U^{(k)} - \eta \nabla_{\Delta U} \mathcal{L}^{(k)}$ \Comment{Eq.~(9)}
\EndFor
\State \Return $\hat{y} = \arg\max_{y \in \mathcal{Y}} p_y\big(U_0 + \Delta U^{(K)}\big)$
\end{algorithmic}
\end{algorithm}

\subsection{Baselines and Evaluation Protocol}
\label{sec:supp_baselines}
\noindent\textbf{AIF}~\cite{liu2026aif} masks a fraction $r$ of the image tokens, selected by the entropy of their attention across layers,
so that no text position can attend to them.
Its released code supports LLaVA-1.5 with a fixed ratio (default $r{=}0.6$);
we port it to Qwen2.5-VL and InternVL and keep its selection function unchanged.
Tab.~\ref{tab:main} reports the best of $r\in\{0.30, 0.45, 0.60, 0.75\}$ in every cell, and Tab.~\ref{tab:supp_fixed} the default $r{=}0.6$.
On V$^*$ at native resolution, AIF with the shared readout stays below the base model at every ratio (at most 78.5\%),
so Tab.~\ref{tab:main} reports it with its own answer protocol (the prompt of its paper with free-text generation), which gives 81.2\%.

\noindent\textbf{TTRV}~\cite{ttrv} adapts the model on unlabeled questions by reinforcement learning before testing.
We use the official code with its released settings (learning rate $5{\times}10^{-7}$, 32 votes, 16 trained responses,
20 adaptation questions, two epochs) and three seeds.
We run two variants: adaptation on 20 questions of the evaluated set, as in its paper, and adaptation on a disjoint fold.
Tab.~\ref{tab:main} reports the best average over variant, epoch and readout in every cell,
and Tab.~\ref{tab:supp_fixed} the disjoint variant after the second epoch.
At native resolution, TTRV runs out of memory on four 80\,GB GPUs.

\noindent\textbf{Crop-based methods.}
Every crop is cut from the original image, capped at 576 tokens and appended to the 576-token input,
and all answers use the shared readout.
\emph{ViCrop}~\cite{zhang2025vicrop}: official code with its relative-attention map, the only variant it provides for Qwen2.5-VL.
The default uses attention layer 17, the official layer of the 3B model scaled to the 28 layers of the 7B model,
and the official 224-px box rule.
ViCrop (tuned) takes the best attention layer (14--27) and crop size
(boxes with six base sizes from 224 to 784\,px and five window sizes between 1/2 and 1/8 of the image side)
on each benchmark, 29--42 settings per benchmark; layer 19 is the best on all three.
\emph{HiDe}~\cite{liu2026hide}: official code with its default setting (layer 15, $\sigma{=}3$, threshold 0.7);
the extra view is its layout-preserving compact image.
\emph{EGG}~\cite{gropl2026egg}: its zoom pipeline, or its first-step map with the crop rules above, whichever is stronger on each benchmark.
\emph{InCue crop} selects the crop with the joint response $J$ at step 0, likewise with its strongest crop rule on each benchmark.
The bottom of Tab.~\ref{tab:crop} and Tab.~\ref{tab:supp_cropgrid} fix the crop rule for all cues.

\noindent\textbf{Issues that the shared protocol avoids.}
The three issues mentioned in Sec.~\ref{sec:experiment} are as follows.
(i) \emph{Chat template.} For InternVL, a chat template without a system turn and without the newline after the role names
leaves many replies empty.
With InternVL3-8B (1 tile), this template and the parser of (ii) give a base accuracy of
47.6\% on V$^*$, 47.9\% on HR-4K and 31.5\% on MME-RW,
against 59.2\%, 57.1\% and 46.4\% when the same TTRV runs use the official template and the option-letter readout.
(ii) \emph{Answer parser.} A parser that takes the last standalone letter A--D of the upper-cased reply as the answer
also counts the article ``a'' as option A.
(iii) \emph{Adaptation samples.} The InternVL3-8B result reported for 20 adaptation samples
coincides with the result for 500 samples.
Our protocol uses the official chat template of each model and the option-letter readout for all methods,
which removes (i) and (ii); TTRV uses 20 adaptation samples.

\end{document}